\documentclass[sn-basic]{sn-jnl}% Basic Springer Nature Reference Style/Chemistry Reference Style

\usepackage{graphicx}%
\usepackage{multirow}%
\usepackage{amsmath,amssymb,amsfonts}%
\usepackage{amsthm}%
\usepackage{mathrsfs}%
\usepackage[title]{appendix}%
\usepackage{xcolor}%
\usepackage{textcomp}%
\usepackage{manyfoot}%
\usepackage{booktabs}%
\usepackage{algorithm}%
\usepackage{algorithmicx}%
\usepackage{algpseudocode}%
\usepackage{listings}%
\usepackage{amssymb}%
\usepackage{pifont}%
\usepackage{array}

\newcommand{\xmark}{\ding{55}}%
\newcommand{\argmin}{\mathop{\mathrm{argmin}}}
\newcolumntype{M}[1]{>{\raggedright\arraybackslash}m{#1}}

\begin{document}

\title[A Complete Survey on Contemporary Few-Shot Learning]{A Complete Survey on Contemporary  Methods, Emerging Paradigms and Hybrid Approaches for Few-Shot Learning}

\author*[1]{\fnm{Georgios} \sur{Tsoumplekas}}\email{gtsoumplekas@metamind.gr}

\author[2]{\fnm{Vladislav} \sur{Li}}\email{v.li@kingston.ac.uk}

\author[1,3]{\fnm{Panagiotis} \sur{Sarigiannidis}}\email{psarigiannidis@uowm.gr}

\author[2]{\fnm{Vasileios} \sur{Argyriou}}\email{vasileios.argyriou@kingston.ac.uk}

\affil[1]{\orgname{MetaMind Innovations P.C.}, \orgaddress{\city{Kozani}, \postcode{50100}, \country{Greece}}}

\affil[2]{\orgdiv{Department of Networks and Digital Media}, \orgname{Kingston University}, \orgaddress{\city{Kingston upon Thames}, \postcode{KT1 2EE Surrey}, \country{UK}}}

\affil[3]{\orgdiv{Department of Electrical and Computer Engineering}, \orgname{University of Western Macedonia}, \orgaddress{\city{Kozani}, \postcode{50100}, \country{Greece}}}

\abstract{Despite the widespread success of deep learning, its intense requirements for vast amounts of data and extensive training make it impractical for various real-world applications where data is scarce. In recent years, Few-Shot Learning (FSL) has emerged as a learning paradigm that aims to address these limitations by leveraging prior knowledge to enable rapid adaptation to novel learning tasks. Due to its properties that highly complement deep learning’s data-intensive needs, FSL has seen significant growth in the past few years. This survey provides a comprehensive overview of both well-established methods as well as recent advancements in the FSL field. The presented taxonomy extends previously proposed ones by incorporating emerging FSL paradigms, such as in-context learning, along with novel categories within the meta-learning paradigm for FSL, including neural processes and probabilistic meta-learning. Furthermore, a holistic overview of FSL is provided by discussing hybrid FSL approaches that extend FSL beyond the typically examined supervised learning setting. The survey also explores FSL’s diverse applications across various domains. Finally, recent trends shaping the field, outstanding challenges, and promising future research directions are discussed.}

\keywords{Few-Shot Learning, Meta-Learning, Transfer Learning, In-Context Learning, Human-Like Learning, Green AI}

\maketitle

\section{Introduction}\label{intro}

In the past decade, there has been significant progress in machine/deep learning fields, mainly due to the abundance of computational resources, the availability of massive datasets, and the development of sophisticated algorithms, like the Transformer architecture \citep{vaswani2017attention}. 
The efficacy of machine and deep learning models has been materialized in diverse fields, including computer vision \citep{he2016deep}, reinforcement learning \citep{silver2017mastering}, natural language processing \citep{brown2020language}, and biomedicine \citep{jumper2021highly}, while also surpassing human performance in some cases \citep{silver2017mastering}. A vital aspect of these models, considered the 'golden rule' of modern deep learning, is their training from scratch on a specific task using a large dataset.

Although modern machine learning models have shown significant success, they still face clear limitations. Due to their increasing size, training these models is a very data- and compute-hungry procedure, while at the same time, these models struggle in continual or incremental learning scenarios without forgetting previously acquired knowledge. As a result, real-world application of these models can be challenging, especially in cases where data is scarce, expensive, or time-consuming to obtain or where there is a lack of necessary computational resources.

On the other hand, one of the most critical aspects of human intelligence is the ability to learn quickly from limited examples and adapt to unfamiliar settings. For instance, young children can identify new object categories given only a few examples. Generally, when learning a new task, an individual uses prior knowledge and experiences and adapts them to the new task based on the provided contextual framework. Consequently, they can adapt more quickly to the new task by transferring already acquired knowledge rather than relearning it from scratch, as machine/deep learning models do.

Parallel to the development of large-scale models, there is a growing interest in integrating such aspects of human-like learning processes within the existing machine learning paradigm. As a result, to bridge the gap between machine and human learning, Few-Shot Learning (FSL) has emerged as a promising direction. More precisely, FSL aims to develop models that can generalize within a specific task, even with a limited number of training samples available. This is typically achieved by using prior knowledge extracted from similar tasks. Notably, such models can adapt to novel tasks with small datasets, which is difficult for conventional machine learning models without the risk of overfitting. Additionally, they can significantly reduce the need for collecting and preprocessing extensive amounts of data for training while also minimizing the time required for adapting to new data. On a more practical level, FSL can also enable learning of rare cases, which is crucial in applications such as drug discovery \citep{altae2017low}.

Leveraging prior knowledge to adapt to novel tasks has been explored since the 1980s via self-referential learning \citep{schmidhuber1987evolutionary} and learning biologically plausible rules that enable adaptation to new tasks \citep{bengio1990learning}. Other early approaches have also included object recognition from a single image using Bayesian models \citep{fei2006one}, learning data augmentations \citep{miller2000learning}, and metric learning \citep{fink2004object}. However, the proliferation of FSL and its influence on various aspects of modern deep learning over the past decade have necessitated consolidating knowledge on recent advancements in the field.

One of the first surveys on FSL was conducted by \cite{wang2020generalizing}, who provided the first formal definition and taxonomy of existing methods and applications in the field. This taxonomy was later extended in \cite{parnami2022learning} to include various hybrid methods beyond the standard supervised setting. More recently, in \cite{song2023comprehensive}, a novel taxonomy has been proposed based on how prior information is abstracted in FSL and the different challenges arising in each case. Meanwhile, other surveys such as in \cite{gharoun2024meta} have narrowed the focus to meta-learning approaches and their application in the field of FSL. Additionally, several recent surveys have focused on specific application areas of FSL, such as FSL in medical imaging \citep{pachetti2024systematic}, FSL in channel state information-based human sensing \citep{wang2024review}, and few-shot object detection \citep{kohler2023few}, among others.

However, there has not yet been a comprehensive overview of FSL that covers the full range of modern FSL methods and applications. Specifically, certain existing surveys have primarily focused on meta-learning approaches for FSL \citep{gharoun2024meta} while most recent attempts to provide a broader perspective on the field have not captured emerging FSL paradigms such as the trending in-context learning \citep{song2023comprehensive, gharoun2024meta}. Meanwhile, discussions regarding the extensions of FSL beyond the supervised learning setting using hybrid methods have been limited in the past \citep{parnami2022learning}. Finally, several surveys \citep{pachetti2024systematic, wang2024review, kohler2023few} have focused exclusively on specific application fields of FSL.

\begin{table}[t]
    \caption{Comparison of existing FSL surveys and our survey. Symbol \checkmark means the topic is covered while \xmark \ means the topic is not covered.}
    \label{table:survey-comparison}
    \centering
    \begin{tabular}{M{1.5cm} M{1.7cm} M{2.4cm} M{2.1cm} M{3.2cm}}
        \toprule
        \textbf{Survey} & \textbf{In-Context Learning} & \textbf{Novel Meta- Learning Model Families} & \textbf{Non Meta- Learning FSL Methods} & \textbf{Hybrid Methods Beyond Supervised FSL} \\
        \midrule
        \cite{wang2020generalizing} & \xmark & \xmark & \checkmark & \xmark \\
        \cite{parnami2022learning} & \xmark & \xmark & \checkmark & \checkmark \\
        \cite{song2023comprehensive} & \xmark & \xmark & \checkmark & \xmark \\
        \cite{gharoun2024meta} & \xmark & \checkmark & \xmark & \xmark \\
        \midrule
        \textbf{Ours} & \checkmark & \checkmark & \checkmark & \checkmark \\
        \bottomrule
    \end{tabular}
\end{table}

This paper aims to address the current lack of a broader overview in the field of FSL by providing a comprehensive and up-to-date survey. In particular, it includes an exploration of novel model families under existing FSL paradigms, such as neural processes and probabilistic meta-learning, incorporates novel FSL paradigms, such as in-context learning, and offers a more detailed review of hybrid FSL methods that extend FSL beyond supervised learning settings. Table \ref{table:survey-comparison} illustrates how our survey integrates these various elements extending upon previous surveys in the field. Finally,  emerging challenges and future directions for FSL in the era of large-scale deep learning models such as human-like learning and Green AI are discussed. Overall, our contributions can be summarized as follows:

\begin{itemize}
    \item We propose a novel taxonomy that builds upon previously proposed ones and extends classic approaches, such as meta-learning, transfer learning, and data augmentation, by introducing new subcategories, including neural processes and probabilistic extensions of meta-learning. We primarily focus on recent advancements while also including well-established methods.
    \item We present, for the first time in an FSL-related survey, an analysis of the recently explored in-context learning within the FSL framework, examining its similarities and differences with more traditional FSL approaches as well as various techniques that combine these two paradigms.
    \item We explore FSL’s connections with relevant learning problems and review various existing hybrid approaches that extend the standard supervised FSL framework to different settings, including unsupervised FSL, few-shot federated learning, cross-domain FSL, and class-incremental FSL, among others.
    \item We highlight the latest trends in FSL and identify some arising challenges that need to be tackled to advance the field. These can bring additional attention to the field of FSL, sparking interest in further advancements.
\end{itemize}

The remainder of this survey is organized as follows. Section \ref{background} provides an overview of FSL, including its formal mathematical formulation, commonly used datasets, and related fields of interest. Section \ref{fsl_models} introduces the proposed taxonomy and provides an overview of various recent and standard approaches within each category. Section \ref{beyond-supervised-FSL} extends the presented taxonomy beyond the supervised FSL setting by introducing hybrid methods at the intersection of FSL and different learning paradigms. Section \ref{fsl_applications} analyzes FSL's fields of application. Section \ref{trends-challeges} analyzes multiple trends that have emerged in recent years and discusses less explored directions within the field that are promising for future research. Finally, Section \ref{conclusion} concludes the survey.

\section{Background}\label{background}

Before delving deeper into the various methods used within the FSL field, it is essential to provide a rigorous mathematical and practical formulation of the problem these methods try to solve. Next, an overview of some standard benchmarks, as well as some emerging ones in novel settings, is provided. Finally, the relationship between FSL and different learning paradigms is examined.

\subsection{Problem Formulation}\label{formulation}

As previously mentioned, FSL refers to a set of learning problems in which machine learning models aim to adapt and be able to generalize given only a limited number of training examples. \cite{wang2020generalizing} provide a more formal definition framing FSL using the same terminology as the one used for defining standard machine learning. In general, given a task \textit{T} and a performance measure \textit{P}, a model is said to learn when its performance \textit{P} on the task \textit{T} is improved by leveraging experience \textit{E} related to the task (e.g., training data).

More specifically, the experience \textit{E} associated with \textit{T} is a dataset $D = \{(x_i, y_i)\}_{i=1}^N$ of \textit{N} samples, where $x_i \in \mathbb{R}^I$, $y_i \in \mathbb{R}^O$ are the input feature vector and corresponding label, respectively. In general, $(x_i, y_i) \sim p(x,y)$, meaning that there is an underlying joint distribution $p$ that generates the data samples within this task. The aim is to create a model $f_{\theta} : \mathbb{R}^I \to \mathbb{R}^O$ parameterized by $\theta$, which is optimized according to a performance measure which typically is a loss function $\mathcal{L} : \mathbb{R}^O\times \mathbb{R}^O \to \mathbb{R}$. This leads to an optimal set of model parameters $\theta^*$ for which:

\begin{equation}
    \theta^* = \argmin_{\theta}\mathbb{E}_{(x,y) \sim p(x,y)}[ \mathcal{L}(f_{\theta}(x), y)]
\end{equation}

Since $p(x, y)$ is unknown, $\theta$ is optimized through \textit{empirical risk minimization} by approximating the expected loss over the joint distribution by the average loss on the training set samples:

\begin{equation} \label{eq_b}
    \theta^* = \argmin_{\theta} \left[ \frac{1}{N} \sum_{i=1}^N \mathcal{L}(f_{\theta}(x_i), y_i) \right]
\end{equation}

Based on this definition, FSL is substantially different from typical supervised machine learning in the sense that there is less training data on which the model can be trained and gain experience for the task. Under these learning conditions, a typical learning model would severely overfit the training dataset and thus could not generalize to unseen examples. As a result, standard FSL approaches rely on extracting knowledge from a set of similar tasks, each consisting of only a few training examples.

Specifically, instead of a single learning task \textit{T}, we now consider a group of tasks $\{T_j\}_{j=1}^M$, which are realizations of a task distribution $q(T)$. The derivation of the tasks from the same distribution implies the existence of a common shared structure that can be extracted and leveraged to adapt more quickly to these tasks. As a result, in this case, the aim is to find an optimal set of parameters $\theta^*$, which minimizes the loss across all given tasks. These optimal parameters would satisfy the following:

\begin{equation} \label{eq_a}
    \theta^* = \argmin_{\theta} \mathbb{E}_{T \sim q(T)} [ \mathbb{E}_{(x,y) \sim p(x,y)} \mathcal{L}(f_{\theta}(x), y)]
\end{equation}

The existence of the two expectations, one over the task distribution and one over the data generating process within each task, in (\ref{eq_a}), imposes an additional level of intractability. As a result, similarly to how the expected error was handled in (\ref{eq_b}), an approximation of the expected error over the task distribution is necessary. Each of the tasks $\{T_j\}_{j=1}^M$ corresponds to a dataset $\{D_j\}_{j=1}^M$ and therefore, the expected loss over all tasks can be approximated by the average loss on the available tasks used during training. For clarity, we adhere to the nomenclature commonly used in meta-learning and FSL and refer to this set of datasets as the \textit{meta-train} set. Additionally, a similar set of datasets from the same task distribution, named \textit{meta-test} set, can be used to evaluate the model on novel tasks.

Accordingly, within each task in the \textit{meta-train} set, a set of data points available during training is used for the error approximation within the task. This set is referred to as the \textit{support set}, while a second set of the task's data points is used for evaluation within the task, named the \textit{query set}. Precisely, each dataset $D_j \in \{D_j\}_{j=1}^M$ is split into $D_j = (D_j^S, D_j^Q)$ where $D_j^S$, $D_j^Q$ are two disjoint sets that correspond to the \textit{support} and \textit{query} sets of the task. Combining all of the above, (\ref{eq_a}) can be approximated by:

\begin{equation} \label{eq_c}
    \theta^* = \argmin_{\theta} \left[ \frac{1}{M} \sum_{j=1}^M \left[ \frac{1}{N} \sum_{(x, y) \in D_j^Q} \mathcal{L}(f_{\theta|D_j^S}(x), y) \right] \right]
\end{equation}

It is worth noting here that $\theta$ is optimized over the \textit{query} set samples of each task. Typically, the model parameters are refined within each task based on the provided \textit{support} set, and the evaluation on the \textit{query} set is performed using these refined parameters. This parameter refinement is visible in (\ref{eq_c}) on the dependence of $\theta$ from the \textit{support} set $D_j^S$ of the examined task. Since the aim of FSL models is to be capable of generalizing within each given task, it makes sense that the loss calculation is performed on samples different than the ones used for refining the model. 

\subsection{N-way K-shot Formulation}

Historically, research around FSL has mainly focused on image classification. In this particular setting, one typically refers to the \textit{N-way-K-shot} classification \citep{vinyals2016matching}, in which each support set contains samples from \textit{N} different classes, with \textit{K} different examples in each class. As a result, each support set consists of $|S| = NK$ training samples. Similar formulations can also be implemented for different learning problems. For instance, in few-shot semantic segmentation, each task might consist of various images of the same object to be segmented, and in few-shot regression, each task corresponds to a different output function with only a few input-output pairs provided in the support set.

\subsection{Datasets} \label{datasets}

Various datasets suitable for FSL training and evaluation have been proposed throughout the years, with \textit{miniImageNet} \citep{vinyals2016matching} and \textit{Omniglot} \citep{lake2015human} being the most widely used for few-shot image classification. In particular, \textit{miniImageNet} consists of 60000 images separated into 100 different classes originating from \textit{ImageNet}, and \textit{Omniglot} contains images of 1623 handwritten characters, with 20 different samples included for each character. Other commonly used datasets include \textit{tieredImageNet} \citep{ren2018meta}, which alleviates \textit{miniImageNet}'s lack of class diversity, having 608 classes grouped in 34 class categories, and \textit{CIFAR-FS} \citep{bertinetto2018meta}. Finally, datasets such as \textit{CUB-200-2011}, \textit{tinyImageNet}, \textit{Aircraft} and \textit{Stanford Cars}, although not initially developed for FSL purposes, have also been used due to their small size and variety of classes.

However, these datasets lack variability in their number of classes and samples per class, and they are also not suitable for evaluating out-of-domain generalization. To combat these issues, datasets such as \textit{Meta-Dataset} \citep{triantafillou2019meta} and \textit{ORBIT} \citep{massiceti2021orbit} have emerged that combine images from different datasets. Extending this framework, there has also been a bloom of other datasets in recent years, aiming to include images from various domains, such as \textit{BSCD-FSL} \citep{guo2020broader}, which includes agriculture, satellite, and dermatology. Other recent approaches include \textit{Meta-Album} \citep{ullah2022meta} focusing on a more realistic cross-domain evaluation of FSL models, spanning across ten distinct domains and \textit{HARD-MD++} \citep{basu2022hard} which is the amalgamation of various tasks considered hard to solve from \textit{Meta-Dataset} and other FSL benchmarks, spanning across four domains. Finally, \textit{Meta Omnium} \citep{bohdal2023meta} extends the FSL framework outside of image classification by providing a benchmark for evaluation of generalization across different learning problems such as key-point localization, semantic segmentation, and regression.

Apart from image classification, various other datasets related to specific problems such as semantic segmentation \citep{li2020fss}, regression \citep{gao2022matters}, pose estimation \citep{xu2022pose}, drug discovery \citep{stanley2021fs} and robotics 
\citep{yu2020meta} have been proposed, however they usually lack the diversity displayed by modern image classification few-shot benchmarks.

\subsection{Relevant Learning Problems} \label{relevant_problems}

Learning efficiently with few data, transferring knowledge across different datasets, and leveraging auxiliary tasks to enhance learning have been well-known research topics within the machine learning community for years. Under these settings, which are particularly challenging for standard supervised learning, various approaches have been proposed throughout the years. In this section, we compare these types of learning with the FSL paradigm, delineating their similarities and differences. Additionally, Section \ref{beyond-supervised-FSL} provides a more detailed discussion, where several hybrid methods that lie at the intersection of FSL and these learning paradigms are presented.

\bmhead{Meta-Learning} The main goal of meta-learning is to enable faster adaptation to novel tasks by utilizing previously acquired knowledge. To achieve this, meta-learning models extract the common structure from a pool of tasks and use it as an inductive bias for rapid adaptation to unseen tasks with scarce data. In general, meta-learning has been widely used to solve FSL tasks (see Section \ref{meta-learning-approaches}). However, its goal of generalizing to novel tasks is not necessarily restricted to the assumption of limited data as in FSL.

\bmhead{Weakly Supervised Learning} This type of learning includes incomplete supervision, where both labeled and unlabeled data is utilized during training, inexact supervision that focuses on training with data that has coarser labels than the ones the model wants to predict, and inaccurate supervision that deals with noisy and incorrect data labels. Although both FSL and weakly supervised learning (WSL) rely on introducing inductive biases by leveraging other data sources, in WSL, these originate from unlabeled data, while in FSL, they usually originate from labeled data from different tasks or pretrained models.

\bmhead{Semi-Supervised Learning} A typical case of weakly-supervised learning is semi-supervised learning, where the model is trained on a few labeled data and a large number of unlabeled data. Typically, the semi-supervised model aims to infer a general structure, acting as an inductive bias, from the unlabeled data and then leverage it to boost its performance on the few labeled data. This bears great similarity with FSL approaches that infer inductive biases from outer sources. However, FSL uses labeled data from other tasks to extract prior knowledge.

\bmhead{Self-Supervised Learning} Similarly to FSL, self-supervised models are initially trained on pretext tasks. However, in self-supervised learning (SSL), these tasks are typically self-generated and consist of unlabeled data, while FSL tasks are labeled and predefined. Additionally, in SSL, there is no restriction on whether these downstream tasks contain only a few data. Generally, the two paradigms have a close relationship \citep{ni2021close} since they aim to extract robust and transferable representations that can be used in downstream tasks. As a result, various recent approaches have incorporated self-supervised techniques such as contrastive learning within the FSL framework (see Section \ref{self_supervised_fsl}).

\bmhead{Transfer Learning} The goal of transfer learning is to transfer knowledge from a source domain where data is widely available to a target domain where data is limited. Regarding FSL, various transfer learning approaches based on model finetuning have successfully been used (see Section \ref{transfer-learning}). However, transfer learning deals with the source (abundant data) and target (limited data) tasks only, and its goal is to transfer knowledge from the former to the latter. On the other hand, FSL deals with multiple tasks, each containing only a few data, and its goal is to extract knowledge from the underlying structure shared among tasks.

\bmhead{Domain Adaptation/Generalization} Domain adaptation aims to train models capable of maintaining a solid performance when evaluated on out-of-distribution samples, while domain generalization extends this framework to create models that exhibit robustness under any distribution shift \citep{khoee2024domain}. Recently, domain adaptation/generalization has been studied within the FSL framework, expanding it from creating models capable of adapting to new tasks with only a few data to also being robust to out-of-distribution tasks and data.
 
\bmhead{Multi-Task Learning} In multi-task learning (MTL), the goal is to train a model that is jointly optimized on a set of tasks. Each task contributes to the optimization of all tasks, and consequently, performance is increased compared to training a separate model for each task. This type of training is similar to FSL, where shared information is extracted from training tasks and is subsequently used to improve performance on novel tasks. However, MTL focuses on a predetermined set of tasks, while FSL aims to generalize to novel tasks not seen during training.

\bmhead{Zero-Shot Learning} The major difference between zero-shot learning (ZSL) and FSL is the number of available training samples within each task. In particular, FSL tasks contain a small number of training samples, while ZSL contains none. As a result, while FSL relies on adapting to these few samples, ZSL leverages external data, usually from different modalities, to enable learning in these tasks. Overall, both approaches introduce inductive biases that facilitate effective learning in novel tasks, yet their sources differ significantly.
 
\bmhead{Continual Learning} In continual learning, the goal is to develop models that can incrementally acquire new knowledge from different tasks without their performance deteriorating due to catastrophic forgetting \citep{khodaee2024knowledge}. While both FSL and continual learning incorporate learning across various tasks, FSL methods are trained offline, while continual learning is heavily based on online learning. This shifts the focus of continual learning towards incremental accumulation of knowledge, while FSL focuses on leveraging prior information to generalize within tasks with a limited number of data efficiently.

\section{Few-Shot Learning Models} \label{fsl_models}

\begin{figure}
 \begin{center}
  \includegraphics[width=0.95\textwidth]{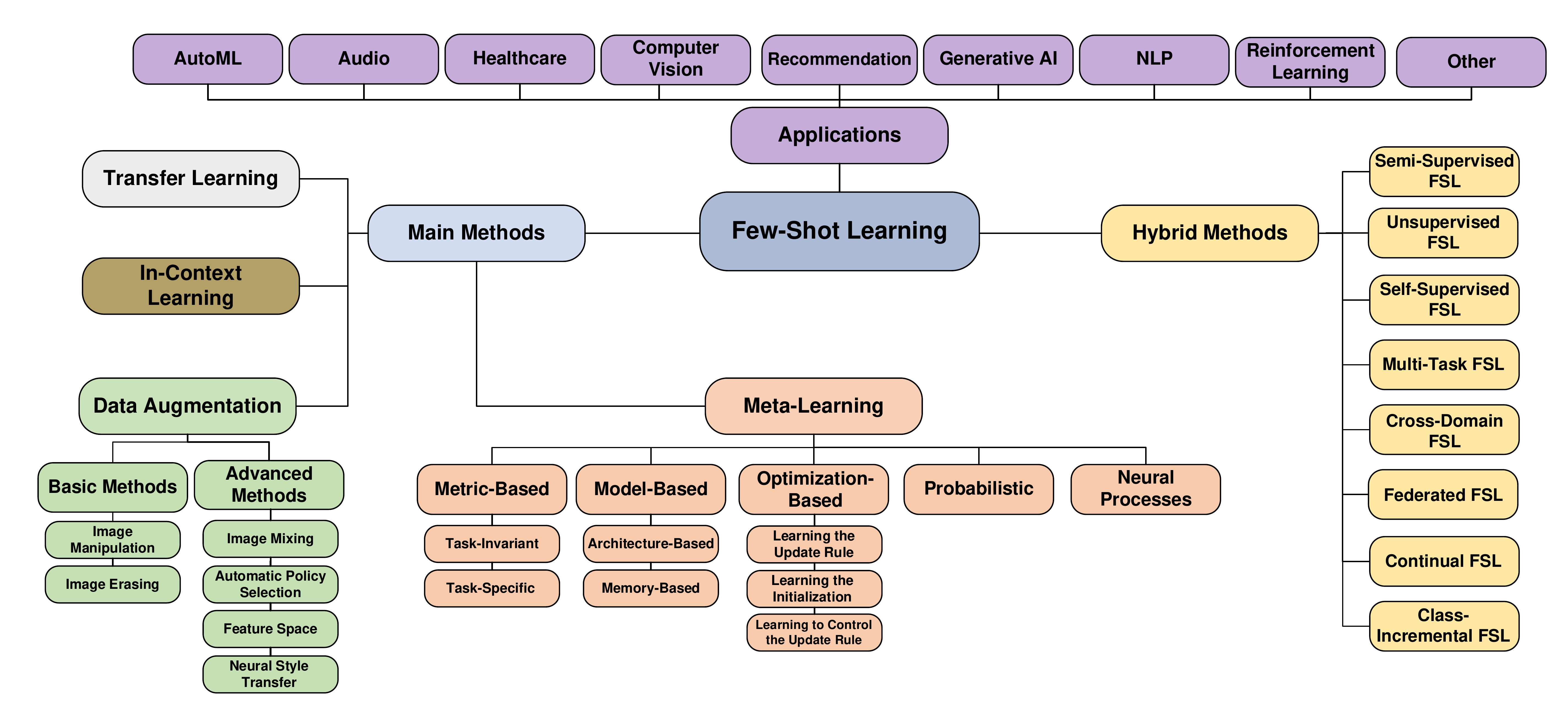}
 \end{center}
 \caption{A taxonomy of FSL methods and their applications. This includes main methods focused on supervised FSL settings (Section \ref{fsl_models}), hybrid methods for FSL beyond the supervised setting (Section \ref{beyond-supervised-FSL}), and FSL fields of application (Section \ref{fsl_applications})}
 \label{img:fsl_methods_tree}
\end{figure}

In this section, we analyze some of the major approaches proposed over the years in the context of FSL. Most existing surveys have mainly focused on \textit{meta-learning}, \textit{transfer learning}, and \textit{data augmentation} techniques. Yet, due to the recent bloom that FSL has seen, we extend this framework to include additional families of methods such as \textit{neural processes} under the meta-learning methodologies, and the emergent \textit{in-context learning} which is commonly found in the recent large language model (LLM) literature. Our focus is on novel approaches that have surfaced in recent years. Yet, an overview of older but well-established methods is also provided to delineate the background upon which more recent works have been developed. Figure \ref{img:fsl_methods_tree} provides an overview of the different types of methods within the FSL paradigm as well as some of their major fields of application.

\subsection{Meta-Learning Approaches} \label{meta-learning-approaches}

Traditionally, meta-learning has been the leading paradigm to deal with FSL. As a result, a wide variety of approaches have been proposed over the years that fall under this category, including \textit{metric-based}, \textit{model-based}, \textit{optimization-based}, and \textit{probabilistic} methods, as well as \textit{neural processes}.

\subsubsection{Metric-Based Methods}

Due to data scarcity in the FSL setting, obtaining accurate and robust decision boundaries for classification tasks is extremely hard, especially for high-dimensional inputs, such as images. The goal of metric-based methods is to learn an embedding space where it is easier to classify the embedded feature representations of each task's data. At the same time, a suitable metric, either predefined or learned, is used to discriminate the different representations. The main component in these approaches is the feature representation extractor, which can be either \textit{task-invariant}, meaning the embedding space is the same for all tasks, or \textit{task-specific}, meaning the embedding space is specifically tailored for each task.

\paragraph{Task-Invariant Representations} 

One of the first FSL models based on metric learning was Siamese Neural Networks \citep{koch2015siamese}. Specifically, their objective is to learn how to discriminate between two images, similar to contrastive learning. Under this setup, both support and query set samples are encoded using the same Convolutional Neural Network (CNN) feature extractor, and classification is done based on the $\ell_1$ distance between each query set sample's representation and the representations of the support set samples.

\begin{figure}
    \centering
    \begin{minipage}{0.46\textwidth}
        \centering
        \includegraphics[width=0.93\textwidth]{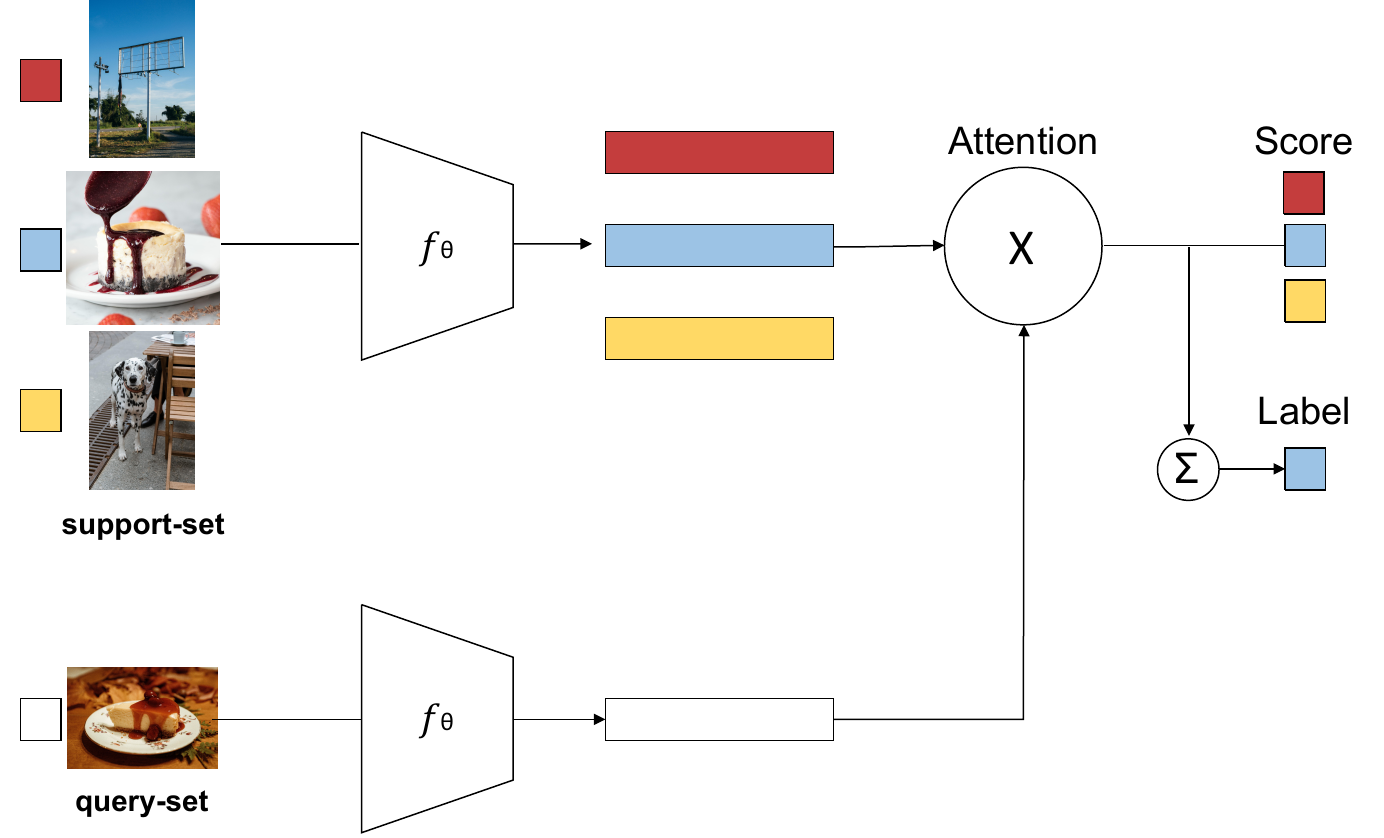}
        \caption{Matching networks architecture \citep{vinyals2016matching}. Predictions are made using an attention mechanism to compare the representations of the support and query set samples}
        \label{img:metric-based-matching-nets}
    \end{minipage}\hfill
    \begin{minipage}{0.52\textwidth}
        \centering
        \includegraphics[width=0.99\textwidth]{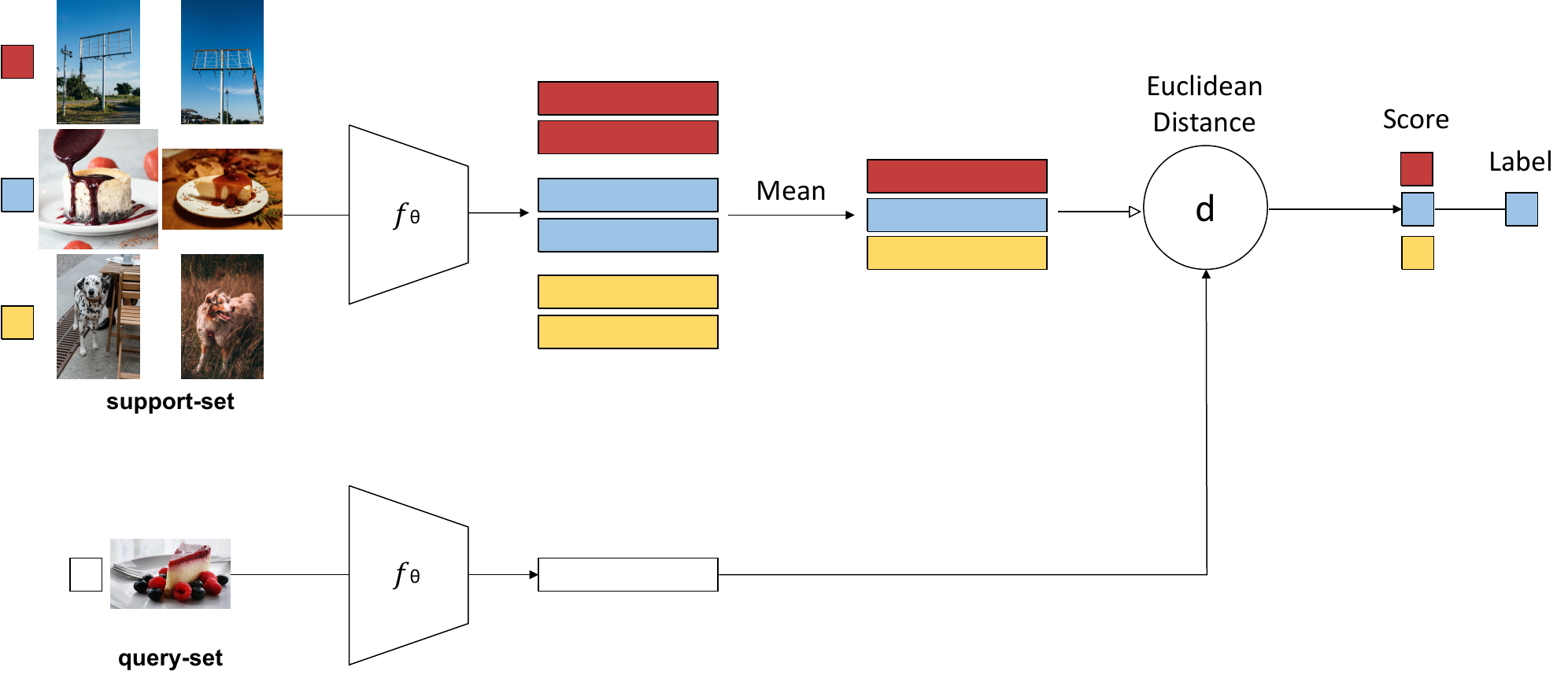}
        \caption{Prototypical networks architecture \citep{snell2017prototypical}. A mean representation is calculated for each class, and the query sample is assigned to the class of the nearest centroid}
        \label{img:metric-based-proto-nets}
    \end{minipage}
\end{figure}

On the other hand, Matching Networks \citep{vinyals2016matching} employ different CNN embedding networks for support and query sets. The predicted output is the weighted sum of the support set sample labels, and the weights are calculated using an attention mechanism based on cosine similarity between the query and support set samples' representations. Full Context Embeddings are also proposed, where support and query set samples are contextualized on the support set embeddings using an LSTM model. Figure \ref{img:metric-based-matching-nets} shows a high-level overview of Matching Networks' architecture.

Another popular method, Prototypical Networks \citep{snell2017prototypical}, relies on learning a suitable feature extractor whose feature representations of the same class samples can be clustered around a single prototype representation. A single CNN feature extractor is used for both support and query set samples, and each class prototype is the mean of the representations belonging to that class. Next, a simple nearest class prototype algorithm is employed to classify each query sample. Figure \ref{img:metric-based-proto-nets} shows a high-level overview of Prototypical Networks' architecture. Recently, Prototypical Networks have been extended in the setting of meta-learning with limited tasks by training them on both real and synthetic tasks. For the creation of synthetic tasks, Conditional Batch Normalization (CNB) is applied to the data of an existing task to create modulated features either using a deterministic objective or via variational inference \citep{sun2023metamodulation}.

\begin{figure}[h]
    \centering
    \begin{minipage}{0.45\textwidth}
        \centering
        \includegraphics[width=0.9\textwidth]{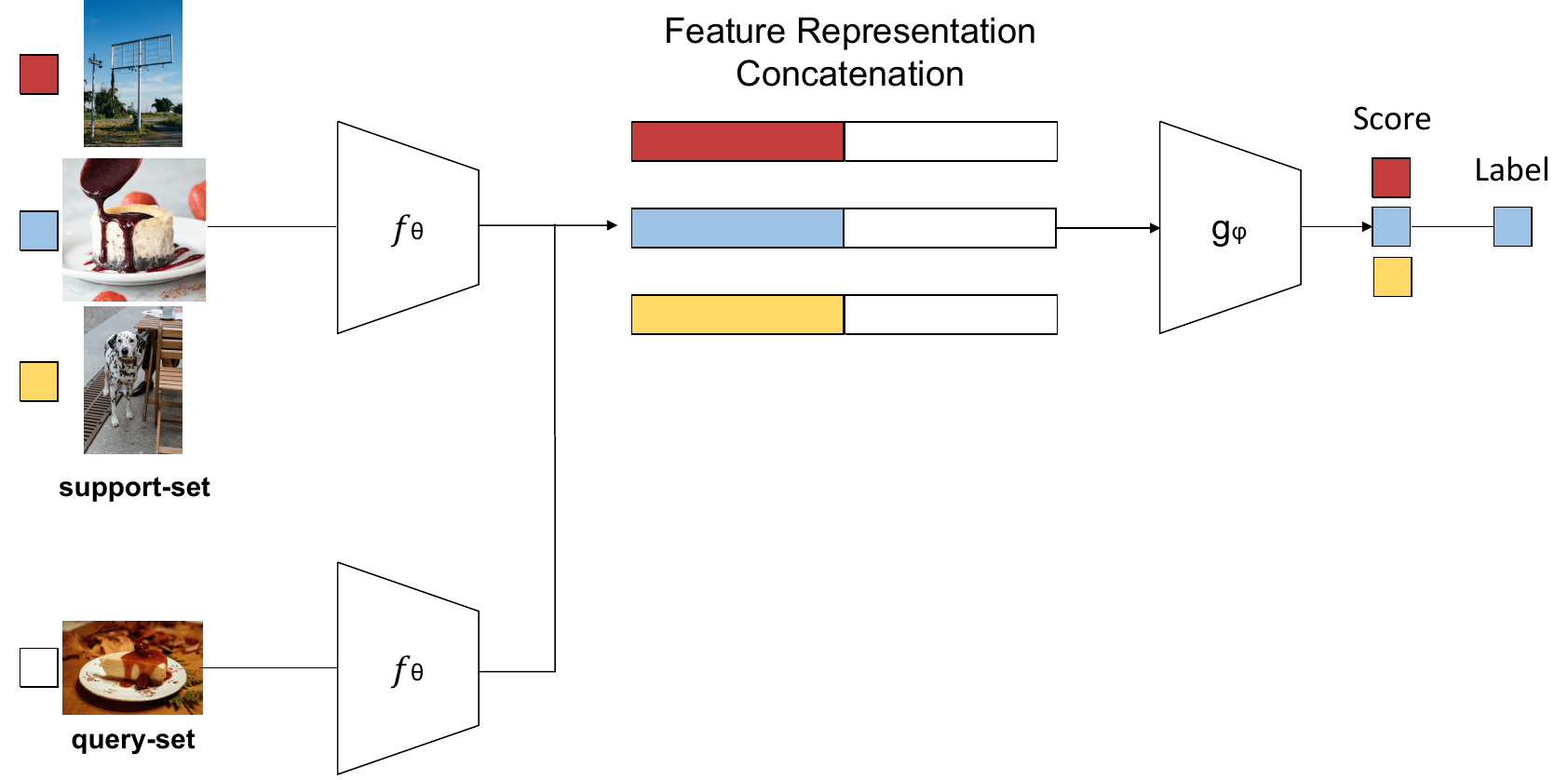}
        \caption{Relation networks architecture \citep{sung2018learning}. Support and query set representations are concatenated and classified based on their similarity following a metric learning approach}
        \label{img:metric-based-relation-nets}
    \end{minipage}\hfill
    \begin{minipage}{0.45\textwidth}
        \centering
        \includegraphics[width=0.9\textwidth]{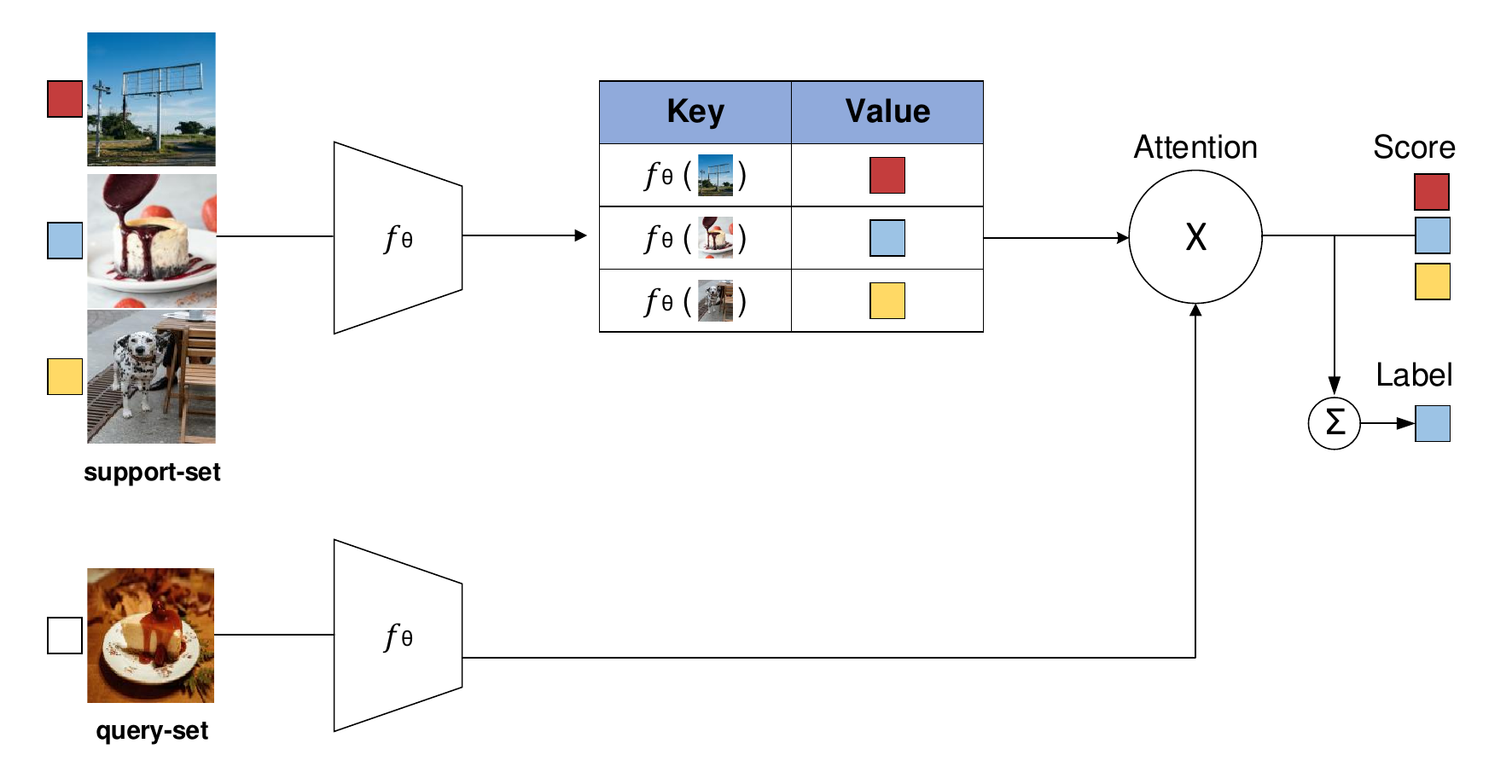}
        \caption{Memory-Augmented Neural Networks architecture \citep{santoro2016meta}. An external memory module is utilized to store and retrieve refined sample representations within each task}
        \label{img:memory-based-MANN}
    \end{minipage}
\end{figure}

Similar to Prototypical Networks, Relation Networks \citep{sung2018learning} also use a single feature extractor for both support and query sets and utilize class prototypes. However, in Relation Networks, the nearest class prototype algorithm is replaced by an MLP module that learns with which prototype the query representation is the most similar. In general, this can be seen as a metric learning approach where, instead of using a predefined distance metric (e.g., Euclidean distance in Prototypical Networks), the metric is learned by a neural network. Figure \ref{img:metric-based-relation-nets} shows a high-level overview of Relation Networks' architecture.

In \cite{garcia2017few}, FSL is approached by a different perspective where each task is modeled as a partially observed graph. While the feature representations of support and query set samples are still extracted to form the graph, a Graph Neural Network is used to draw inferences, acting as a form of learnable metric. Notably, the proposed method is general enough and includes previous architectures, such as Prototypical Networks and Siamese Networks, as special instances. It can also be applied for semi-supervised and active FSL.

Finally, DeepEMD \citep{zhang2020deepemd} aims to address the limitations of using global features by leveraging local feature characteristics that provide more discriminative and transferable information across categories. Specifically, instead of comparing representations of whole images, these are broken down into patches, and Earth Mover's Distance (EMD), which is incorporated as a layer, is used to calculate their similarity. Additionally, a cross-reference mechanism is used to discard patches that introduce noise due to cluttered backgrounds and large intra-class appearance variations.

\paragraph{Task-Specific Representations} 

One of the first works to address the limitations of using task-independent feature extractors was TADAM \citep{oreshkin2018tadam}, which introduced the use of an extractor contextualized on each task's data. Contrary to Relation Networks' representation contextualization that is done as a post-processing step, in TADAM, the feature extractor is directly modulated via learnable task-specific shift and scale vectors for each convolutional layer, and a learnable scale parameter is introduced in the model's loss function. Finally, metric scaling is proposed to bridge the gap between using different metrics.

FEAT \citep{ye2020few} also leverages task-specific representations by contextualizing each support set sample to the rest of the set's samples using a self-attention mechanism, while a contrastive objective is incorporated into the loss function to enhance discrimination between different classes. A transformer-based architecture has also been used in Sparse Spatial Transformers \citep{chen2023sparse} (SSFormers). In particular, SSFormers aim to improve DeepEMD by contextualizing an image's local features to the rest of the image, as well as patches belonging to other support set images, by using sparse attention maps of the extracted task-relevant patches.

Recently, various task-specific adaptations of existing task-invariant methods have also been proposed. For instance, finetuning of Prototypical Networks' backbone feature extractor for each task has been proposed, based on the observation that learned feature representation may fail to generalize to novel tasks due to possible distributional shifts \citep{hu2022pushing}. Additionally, in \cite{kim2023universal}, a model that can adapt to different dense prediction tasks by matching similar image patches and using the weighted mean of the output patches as the prediction is proposed. This model can be seen as a generalization of Matching Networks where matching is performed on the latent representations of the labels derived from similar image patches.

Finally, in \cite{hu2022suppressing}, the problem of few-shot semantic segmentation is addressed by learning a feature extractor that addresses different sources of heterogeneity. In particular, it incorporates a cross-sample attention module, the fusion of information between different background patches, and a masked image segmentation scheme to reinforce the capacity of contextual inference. Contrary to previously mentioned approaches, this method focuses on learning more separable feature representations by enhancing intra-class compactness and leveraging their distances during training.

\subsubsection{Model-Based Methods}

While metric-based meta-learning methods focus on producing easily separable feature representation, model-based methods are generally based on developing specific model architectures that enable rapid adaptation to novel tasks. Yet, despite their differences, both methods aim to constrain a model's hypothesis space to facilitate its adaptation using less data. Model-based methods can be either \textit{architecture-based}, where the proposed model leverages an internal memory or a specifically designed combination of components that enable FSL, or \textit{memory-based} methods, where an external memory module is used to retain information related to rare class samples.

\paragraph{Architecture-Based Methods} Some of the earliest works in this line of research have been primarily based on the fact that the hidden states of Recurrent Neural Networks (RNNs) can be utilized to model learning algorithms. For instance, in \cite{duan2016rl}, learning a reinforcement learning (RL) algorithm capable of rapidly adapting to novel tasks is also formulated as an RL problem. Based on the concept that in meta-learning, knowledge acquisition takes place in two different time scales, a GRU network's internal state is used as a fast learner to adapt to a new task with few samples, while Trust Region Policy Optimization is used to optimize the slow learner's policy across tasks. Consequently, the model can generalize within different tasks like multi-armed bandits, tabular Markov Decision Processes (MDPs), and visual navigation (partially observable MDPs). On the other hand, in \cite{wang2016learning}, a similar approach based on RL is applied for structured task distributions. The proposed deep meta-RL approach trains an RNN to act as a learning algorithm that can be applied to different MDPs from the same distribution. Using reward information from various tasks, the RNN can quickly adjust to new tasks during testing without finetuning due to its internal state and ability to model different learning algorithms.

On a more general approach, SNAIL \citep{mishra2017simple} can be applied to different types of problems without any assumptions for the task or the optimization procedure. It uses a generic architecture based on three building blocks: dense blocks (temporal 1D causal convolutions), TCBlocks (dense blocks with increasing dilation rates), and a self-attention block. Interestingly, using temporal convolutions interleaved with attention enables selecting relevant information over a larger context window. Overall, this approach allows for learning an optimal strategy without any built-in assumptions, as in gradient-based meta-learning. 

Finally, some recent approaches have introduced architecture-based models suitable for or inspired by different domains, such as a meta-learning model for forecasting dynamical systems that generalizes across different time series sharing a common structure \citep{wang2022meta}. This is achieved by extracting time-invariant features of each time series dynamics using a CNN-based encoder and then adapting to the dynamics of each specific task via adaptive modules. On the other hand, inspired by the Lottery Ticket Hypothesis \citep{frankle2018lottery}, a gradient-based meta-learning method is proposed in \cite{chijiwa2022meta} to meta-learn a sparse neural network structure suitable for FSL. Specifically, the method aims to learn a parameter mask to derive a sparse subnetwork that demonstrates strong meta-generalization capabilities. This method can be seen as a generalization for architecture-based modeling from hand-crafted architectures to learnable ones.

\paragraph{Memory-Based Methods} External memory modules can store different types of content that facilitate rapid adaptation, such as model weights or sample representations. Meta Networks \citep{munkhdalai2017meta} is a model that uses external memory modules to store fast weights for rapid adaptation across tasks. The model consists of a meta-learner (learning across tasks) and a base-learner (task-specific learning) that are both augmented using additional parameters (fast weights) generated on the fly. These fast weights are used to parameterize each module to rapidly adapt to specific tasks and specific task samples. An external memory module is also used to store these fast weights and retrieve their optimal combination based on the task's input representations.

On the other hand, most memory-based models typically rely on external memories to store refined sample representations that enable fast adaptation to classes with few samples. In \cite{santoro2016meta}, the stored values are retrieved using soft attention and writing to memory is controlled by a content-based memory writer that writes memories to either the least used or the most recently used memory location. Figure \ref{img:memory-based-MANN} shows a high-level overview of the model's architecture. Additionally, in \cite{kaiser2017learning}, an independent memory module is proposed that allows neural networks to achieve one-shot and life-long learning. Specifically, it uses a memory that stores class labels, and the main network's layer activations are used to query the memory based on their cosine similarity with the memory keys. An approximate nearest neighbors algorithm is used to scale the memory, and the memory can be combined with various network types, leading to improved classification of rare samples.

Recently, memory-augmented networks have also been utilized in different applications where rare classes are common. In \cite{gkanatsios2023analogy}, a memory-augmented 3D object scene segmentation network is proposed. The model consists of a retriever that finds the most relevant representations in the memory and a modulator where the retrieved memories are encoded and enhanced by scene-agnostic queries, while attention modules are also used for contextualization of the query images. Finally, in \cite{schimunek2023context}, a memory-enhanced model for few-shot drug discovery is proposed. Specifically, it uses a context module based on Modern Hopfield Networks (MHNs) to extract representations of the support set and the query set conditioned on a large external context set of molecules via an attention mechanism. Attention is also used to contextualize the support and query sets, and overall robustness is enhanced by reinforcing the covariance structure and reducing the noise.

\subsubsection{Optimization-Based Methods}

Optimization algorithms are a crucial component of deep learning. However, they lack guarantees for nonconvex optimization problems like deep neural network optimization and may require many iterations for convergence, leading to suboptimal solutions. Additionally, choosing the best optimizer for different problems is a laborious task. Optimization-based meta-learning aims to learn optimization procedures suitable for optimization across different tasks without multiple iterations or large volumes of data by casting it as a 'learning how to optimize' problem. Subsequently, optimization-based methods can be categorized into three types: learning the \textit{update rule}, learning the \textit{initial parameters}, and learning to \textit{control the update rule}.

\paragraph{Learning the update rule}

Based on the principle that learned features are better than hand-crafted ones, in this setting, using hand-crafted update rules such as Stochastic Gradient Descent (SGD) is replaced by learned rules. This is achieved by casting the design of the rule as a learning problem. One of the first approaches towards that direction has used LSTMs to output the updates of the gradient descent step used within each task \citep{andrychowicz2016learning}. To control all steps of the optimization procedure, an objective dependent on the entire trajectory of learned optimization is used. LSTMs are suitable for this since they maintain past information in their internal state, allowing for previous updates to inform subsequent ones. Meta-Learner LSTM \citep{ravi2016optimization} extends this framework to the FSL setting by learning both the base learner's initialization and update strategy using only a small, predefined number of steps. Specifically, each LSTM cell state represents the parameters of the learner, and the cell state variations correspond to model updates. However, this method does not scale well with the base learner's size, and both methods fail to capture any dependencies between different model parameters. Finally, finding an optimal optimization algorithm has also been cast as an RL problem \citep{li2016learning}. Specifically, learning the optimal way to obtain update steps can be seen as selecting an optimal update policy, which is achieved via guided policy search. A specialized cost function is also used that encourages learning update rules that converge fast.

\paragraph{Learning the initial parameters}

To address hand-designed optimization algorithms' slow convergence, some optimization-based algorithms focus on learning initial parameters that converge to optimal parameters with only a few gradient descent iterations. Model-Agnostic Meta-Learning \citep{finn2017model} (MAML) is the first work that aims to learn a model parameter initialization that allows for rapid adaptation to each given task. In particular, meta-learning is cast as a bi-level optimization problem with two learning loops in two different levels. In the inner loop, the base learner is responsible for adapting the initial parameters in the task's support set and evaluating them on the task's query set. In the outer loop, the meta-learner optimizes the initial model parameters using the query set errors across training tasks. As a result, the shared information across tasks is encapsulated in the model's learned initial parameters, which act as a prior that obviates the need for many optimization steps within each task.

Following MAML, many different extensions and variants have been proposed, aiming to address its limitations. For instance, MAML++ \citep{antoniou2018train} is such an extension that introduces various novel features that aim to enhance MAML's performance and ensure its training stability, such as an outer-loop loss that is dependent on all of the inner loop steps' losses, improved batch normalization for meta-learning, and learned inner loop learning rates. The difficulty of choosing optimal learning rates for MAML due to its training instability is also addressed in \cite{behl2019alpha}, where the authors propose to learn inner- and outer-loop learning rates using hypergradient descent, which adaptively updates the learning rates through gradient descent. Meanwhile, in \cite{rajeswaran2019meta}, an implicit differentiation approach that removes the need for differentiating through the inner loop's optimization path is proposed to address the high computational cost and memory burden imposed by using higher-order derivatives in MAML. Finally, REPTILE \citep{nichol2018first} introduces a first-order approximation of MAML that obviates the need to backpropagate through inner loop gradients. Consequently, this mitigates the need to calculate second-order derivatives and significantly reduces computational complexity.

Multiple approaches have also focused on improving MAML's performance and extending its capabilities. In \cite{rusu2018meta}, the proposed method decouples optimization-based meta-learning from the high-dimensional space of the model parameters by embedding the model parameters to a lower dimensional latent space. The produced decoded initial parameters are specific for each task and conditioned in its support set, and MAML uses the latent variables as input. Lastly, by formulating tasks as discriminating between in-distribution and out-of-distribution samples for a specific class, MAML has also been extended in detecting out-of-distribution samples \citep{jeong2020ood}.

The recent advances in theoretical aspects of meta-learning have also led to the development of various theory-inspired MAML adaptations. In \cite{abbas2022sharp}, to avoid the sharp local minima of MAML's loss function, sharpness-aware minimization, a method that simultaneously minimizes the loss value and the loss sharpness, is integrated into MAML. Additionally, in \cite{wang2023improving}, to improve both meta-generalization and generalization performance within each task after adaptation, standard regularization is applied in the outer loop, and inverted regularization is applied in the inner loop, increasing adaptation difficulty within each task and forcing the model to learn better-generalized hypotheses at the meta-level.

\paragraph{Learning to control the update rule}

Learning the update rule in FSL can lead to update rules with non-converging behavior and lacking the inductive bias of gradient descent. To address these limitations, a branch of optimization-based approaches focuses on learning how to control the updates of existing optimizers through a learnable component that modulates the loss gradient (gradient preconditioning). One of the first approaches in that direction extends MAML by additionally meta-learning the learning rates and learning directions for each parameter of the base learner \citep{li2017meta}. This allows for more expressive learning but doubles the amount of meta-learned parameters, resulting in higher computational complexity. Gradient preconditioning can also take the form of learning which parameters are shared across tasks and which ones are task-specific by using a learnable mask and transforming the model parameters using meta-learned transformation parameters \citep{lee2018gradient}. This can be seen as extending MAML to learn a parameter subspace for task-specific adaptation using a meta-learned distance metric. In \cite{park2019meta}, MAML is extended to meta-learn curvatures. These curvatures are used to transform gradients before passing through the optimizers, thus constituting a form of preconditioning. Preconditioning can also be induced implicitly by introducing nonlinear layers interleaved with the base learner's layers \citep{flennerhag2019meta}. In this case, the occurring preconditioning matrix is meta-learned without backpropagation over the base learner's parameters, leading to increased efficiency and better scaling. Finally, using a preconditioner in MAML's inner loop to enable more efficient learning by reshaping the geometry of the parameter space has also been proposed \citep{kang2023meta}. The preconditioner is task-specific and formulated as a Riemannian metric, equivalent to natural gradient descent, and it alleviates meta-overfitting by relying more on individual task characteristics.

\subsubsection{Probabilistic Extensions} \label{prob_extensions}

To alleviate deterministic meta-learning's inability to quantify uncertainty, various probabilistic approaches have been developed to improve performance and uncertainty quantification by modeling parameters as random variables and obtaining their distributions. In this direction, Bayesian meta-learning is a popular paradigm that fits expressive generative models to small datasets by incorporating inductive priors learned from related tasks. One of the first works to explore probabilistic FSL is in \cite{edwards2016towards}, where a hierarchical version of Variational Autoencoders (VAEs) is proposed that can be used for FSL by learning the statistics of the training data and then using them to classify new samples. The basic idea is to perform variational inference at two levels using a double variational bound where a higher-level latent variable describes datasets and a lower-level latent variable describes individual examples.

Additionally, many subsequent works have focused on extending MAML to the probabilistic setting. In \cite{grant2018recasting}, MAML is reinterpreted as an inference for the parameters of a prior distribution in a hierarchical Bayesian model, and it is shown that the choice of update rule in the inner loop corresponds to a choice of prior over task-specific parameters. To mitigate MAML's inaccurate point approximation, a Bayesian variant of MAML using the Laplace approximation is also proposed. In \cite{finn2018probabilistic}, MAML is extended by learning the joint posterior of initial and task-specific parameters, conditioned on each task's support set, using structured variational inference. An amortized variational inference technique is also proposed to avoid storing separate variational distributions for each task's initial and task-specific parameters. A Bayesian variant of MAML leveraging non-parametric variational inference has also been proposed in \cite{yoon2018bayesian}, which learns the posterior distributions of task-specific parameters using Stein Variational Gradient Descent to approximate the posterior over the weights in the final layer with a particle ensemble. However, the posterior's expressiveness depends on the number of particles in the ensemble, leading to poor scaling regarding memory and computation costs.

Apart from probabilistic MAML extensions, novel architectures have also been proposed, such as deep kernels where both feature extractor and base kernel parameters are meta-learned across tasks \citep{patacchiola2020bayesian}. The deep kernel's parameters are learned in the outer loop by maximizing the marginal likelihood across all tasks using a Gaussian process approach, obviating the need for an inner loop and leading to an efficient and straightforward closed-form solution. Finally, a novel method is introduced in \cite{pavasovic2022mars} that meta-learns a Bayesian prior on the function space to avoid overconfident estimations. The problem is formulated as a stochastic process with different realizations, and the prior's score function is estimated using a transformer encoder architecture.

Lastly, probabilistic meta-learning can also mitigate some of metric-based meta-learning's limitations. Specifically, metric-based meta-learning methods typically obtain a single point estimate for each class, leading to biased estimates due to a lack of data. However, a proposed extension in \cite{zhang2019variational} models each class with its distribution through variational inference, allowing for the calculation of confidence and more accurate classification of novel samples. Expanding on this, in \cite{sun2021amortized}, computational complexity is reduced by using fewer random variables for latent class prototypes. A global model is used to learn the posterior distributions of these variables through amortized inference, and task-specific estimates of variational parameters are obtained using initial parameters learned across tasks.

\subsubsection{Neural Processes} \label{neural_processes}

Parallel to the development of probabilistic models that extend the classic meta-learning framework, significant research effort has revolved around Neural Processes (NPs), another family of probabilistic models suitable for FSL. On a higher level, NPs aim to bridge the gap between deep neural networks (DNNs) and Gaussian Processes (GPs). DNNs excel at function approximation, yet they require retraining for each new function. On the other hand, GPs can quickly infer new functions at test time, but they are computationally expensive and sensitive to the selection of priors \citep{garnelo2018conditional}. NPs can make accurate predictions given a few training data points and scale to complex functions and large datasets. NPs were initially developed as function estimators, suitable for modeling stochastic processes. However, their ability to model different functions using only a few samples makes them ideal for few-shot regression, where each function is a different task, and the context data constitutes the support set.

The first proposed model of this family has been Conditional Neural Processes \citep{garnelo2018conditional} (CNPs) that uses a permutation-invariant neural network to parametrize its dependence on the observations. For each task, an aggregated mean representation of the support set is obtained and used by a decoder that models the distribution parameters of the examined target point. The model is trained end-to-end through gradient descent to maximize the conditional likelihood of a random subset of targets given a random observation set. Extending CNPs, NPs \citep{garnelo2018neural} introduce a more generalized approach by using a global latent variable that allows for better uncertainty modeling. Similar to CNPs, an aggregated mean representation of the support set data is obtained, which conditions the global latent variable. Finally, the decoder receives the target data and a sampled value from the conditioned global latent variable as input and predicts the output values.

Yet, despite their desired properties, these methods lack expressivity power and suffer from underfitting in complex settings. To address these limitations, Attentive Neural Processes \citep{kim2019attentive} (ANPs) introduce an attention mechanism to enhance model performance. Specifically, self-attention is used to capture the interactions between support set samples, and cross-attention is used to obtain the aggregated representation. However, these attention modules also lead to increased computational complexity. Convolutional Conditional Neural Processes \citep{gordon2019convolutional} (CCNPs) have also been proposed as an extension of NPs, which introduces translation equivariance as an inductive bias using a convolutional extension of Deep Sets. Additionally, to enable the modeling of temporal dynamics in sequential data, a combination of RNNs and ANPs has been proposed in \cite{qin2019recurrent}, where the hidden states of an LSTM are used to extract the feature representations. Finally, some recent extensions have also included NPs based on the Transformer architecture \citep{nguyen2022transformer}.

\subsection{Transfer Learning Approaches} \label{transfer-learning}

\begin{figure}
 \begin{center}
  \includegraphics[width=0.8\textwidth]{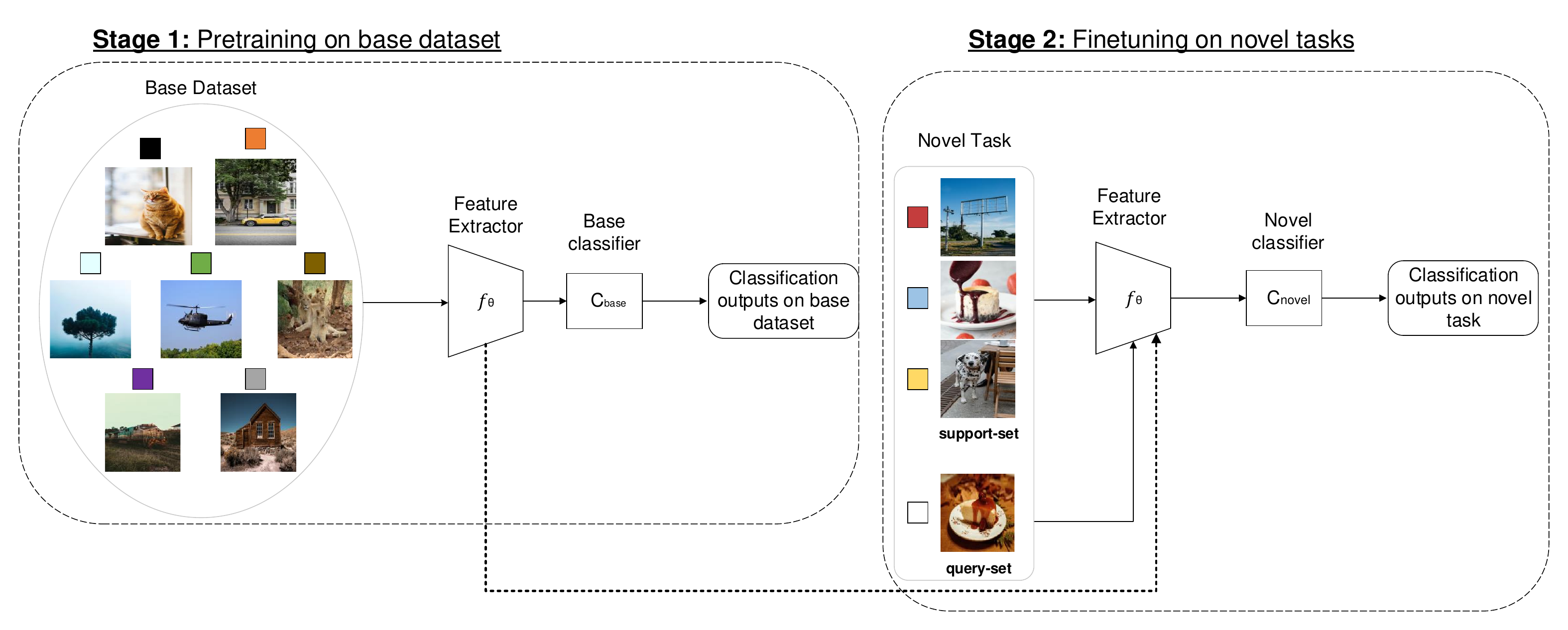}
 \end{center}
 \caption{Two-stage model training in transfer learning. In the first stage, the model is pretrained on a large annotated dataset (base dataset), while in the second stage, it undergoes further finetuning using a smaller target dataset}
 \label{img:transfer-learning}
\end{figure}

Although meta-learning approaches have been successfully used in the context of FSL, they are notorious for their architectural and computational complexity. However, dealing with these types of complexity is neither desirable nor affordable in various real-world settings. In contrast, transfer learning approaches are more straightforward and feasible for such cases, and many novel methods tailored for FSL have been proposed in recent years. In general, most of these methods rely on a two-stage learning procedure that includes pretraining a feature extractor backbone on a large base dataset and then performing finetuning in downstream novel tasks with few samples. Typically, most of the proposed approaches differ in the finetuning stage, which can either include finetuning the whole model or only parts of it. Figure \ref{img:transfer-learning} illustrates a high-level overview of the two-stage procedure.

One of the first works to explore the efficiency of finetuning techniques for FSL has been in \cite{chen2019closer}, where it is argued that finetuning methods' performance has been underestimated in the past. Consequently, a simple baseline that utilizes a pretrained feature extractor backbone and finetunes only the classification head within each few-shot task is proposed. Classification is then performed based on the cosine similarity between the prototypes extracted from the weight matrix of the classifier head and the feature representations of the given samples. Similarly, in SimpleShot \citep{wang2019simpleshot}, a pretrained frozen backbone is also used to extract feature representations. However, a series of simple transformations such as centering and $\ell_2$-normalization are applied to the extracted representations, and the final classification is performed using the nearest neighbor/centroid algorithm.

On a different approach, in \cite{wang2022revisit}, it is suggested that transferring knowledge from a base dataset to new tasks without retraining the pretrained backbone could lead to biased results due to the extraction of distorted features for out-of-distribution samples. To avoid this and prevent overfitting of the feature extractor due to the small number of samples, the whole model is finetuned using an inverted Firth Bias regularizer that regularizes the distribution of the predicted labels to be more uniform, similar to label smoothing. Full model finetuning is also proposed in \cite{dhillon2019baseline}. In particular, the pretrained backbone is combined with a linear classifier head, and a transductive objective is introduced to finetune both the backbone and the classifier within each task, granting it the ability to adapt to different numbers of ways and shots without retraining.

To address both limitations of feature distortion using a frozen pretrained backbone and the possibility of overfitting when finetuning the whole model, partial finetuning, e.g., finetuning only specific layers of the backbone network, has been proposed in \cite{shen2021partial}. To determine the finetuned layers, a genetic programming algorithm is used to learn each layer's learning rates during finetuning, automating the whole procedure. On the other hand, in \cite{shysheya2022fit}, FiLM modules are introduced to modulate the backbone network without retraining all of its parameters, substantially reducing the number of updated parameters. On top of it, a Naive Bayes classifier is used for prediction.

Finally, various other approaches have also emerged, aiming to tackle FSL via model finetuning. In \cite{rizve2021exploring}, the effect of invariance and equivariance is explored in the context of learning better feature representation of samples for FSL. Subsequently, the authors propose learning two additional modules alongside the classification head for enforcing equivariance and invariance via self-supervised learning, and the final model is obtained via self-distillation. Transfer learning has also been combined with meta-learning to bridge the gap between the two methods. Specifically, in \cite{demirel2023meta}, meta-learning is used to tune an object detection model's hyperparameters, while finetuning is used to tackle the few-shot object detection objective in each task.

Overall, finetuning methods can be a great asset in few-shot applications where simplicity and efficiency are crucial. However, there is no clear consensus on whether using a frozen backbone or finetuning it either fully or partially leads to better performance \citep{luo2023closer}. Additionally, pretraining the backbone network in a large base dataset is not always possible in some domains. Finally, cautiousness is needed when there is a domain gap between the base dataset and the novel tasks since it could hinder learning due to negative transfer.

\subsection{Data Augmentation Approaches}

Since FSL is characterized by the lack of adequate data to enable accurate adaptation to the problem at hand, a straightforward solution is to augment the given dataset with new samples. This is the main functionality of Data Augmentation (DA) methods, which reduce overfitting in machine learning models either by enforcing a form of regularization through the existing data or by increasing the number of available samples. In general, DA techniques can further be categorized into basic and advanced DA methods. It is worth noting that FSL has primarily been examined for visual domain tasks, and most of the techniques discussed in this context refer to image input.

\subsubsection{Basic Data Augmentation}

Basic DA methods include fundamental methods widely used in both FSL as well as standard machine learning and can further be categorized into two groups: (a) \textit{Image Manipulation}, and (b) \textit{Image Erasing}. In general, these methods have been widely adopted due to their simplicity, which makes them easy to implement and adds minimal overhead to the training procedure.

\paragraph{Image Manipulation}

This group of DA methods involves altering the visual representation of an image. Some widely used image manipulation techniques include rotation, translation, shearing, cropping, and color jittering. By introducing these geometric and visual modifications, models trained using such data can become more robust by learning feature representations invariant to these alterations. However, hand-picking these transformations beforehand is a challenging task, and simply trying all available methods can add up to a significant computational cost. Consequently, while these methods can be helpful, they can only provide a partial solution in the context of FSL.

\paragraph{Image Erasing}

The main logic behind these approaches is to regularize the model by preventing it from overfitting to particular regions of an image and ignoring other parts. In general, this is achieved by removing specific parts of an image and replacing them with new ones, forcing the model to focus on different features. For instance, Cutout randomly removes square regions (cutouts) of the training images, while Random Erasing randomly selects and erases rectangular regions within an image and replaces them with random or predefined pixel values. On the other hand, a more recent approach, Context-Guided Augmentation \citep{chen2023using}, adds new instances of classes strategically using contextual information to simulate occlusion occurrences, with each segmented instance being selectively positioned onto the object's bounding box in a randomized manner. Overall, image erasing methods can be seen as a form of spatial dropout that induces robustness against object occlusions and different spatial configurations. However, they suffer from the same limitations as image manipulation methods since they must be hand-picked beforehand.

\subsubsection{Advanced Data Augmentation}

The rapid expansion of the computer vision field has led to the emergence of novel sophisticated DA approaches that aim to overcome the limitations of basic DA methods. Advanced DA techniques typically fall into categories such as image mixing, automatic policy selection, feature space augmentations, and neural style transfer.

\paragraph{Image Mixing}

These methods aim to enhance generalization performance across diverse datasets and mitigate the risk of overfitting by integrating information from multiple sources. In Mixup \citep{zhang2017mixup}, this is achieved by blending two or more images by taking a convex combination of their pixel values and their corresponding one-hot label encodings to create novel hybrid samples. Notably, these linear interpolations introduce diversity to the training data and assist models in learning more robust and generalized features with minimal computational overhead. CutMix \citep{yun2019cutmix}, on the other hand, extends this concept by cutting and pasting rectangular patches from different images, fostering spatially coherent variations. More recently, Multi-stage Augmented Mixup \citep{liang2023miamix} (MiaMix), a novel mixing ratio sampling method, has been proposed, consisting of four stages: random sample pairing, sampling of mixing methods and ratios, mixing mask generation and augmentation, and obtaining the mixed output. MiaMix selects a mixing method from a pool of methods and sets the mixing ratio for each mask. The masks are then augmented and mixed in the final stage. Finally, PatchMix \citep{xu2022patchmix} has been proposed to address the existence of biases, such as spurious correlations, in FSL. Specifically, it disentangles causal and non-causal features to improve model invariance and generalizability by augmenting original query images with images from another query while preserving the label of the additional query.

\paragraph{Automatic Augmentation Selection Policies}

Although hand-crafted DA methods have been invaluable in dealing with data scarcity and creating models invariant to various transformations, manually selecting the optimal augmentations to apply is a challenging task that often leads to suboptimal results. As a result, novel approaches have been suggested to automate this selection process by systematically finding the optimal selection policy. AutoAugment \citep{cubuk2019autoaugment} formulates a search space within which each policy consists of multiple sub-policies that encapsulate different image transformations. In this setting, the most effective policy is identified through a search algorithm that maximizes validation accuracy on a target dataset based on RL principles. On the other hand, Randaugment \citep{cubuk2020randaugment} automates this selection by introducing a randomized approach. In particular, it dynamically selects a set of augmentation operations from a transformations pool and applies them with random magnitudes to each image. This approach strikes a balance between variability and consistency while it also obviates using an RL algorithm to find the optimal policy.

\begin{figure}[h]
    \centering
    \begin{minipage}{0.48\textwidth}
        \centering
        \includegraphics[width=0.9\textwidth]{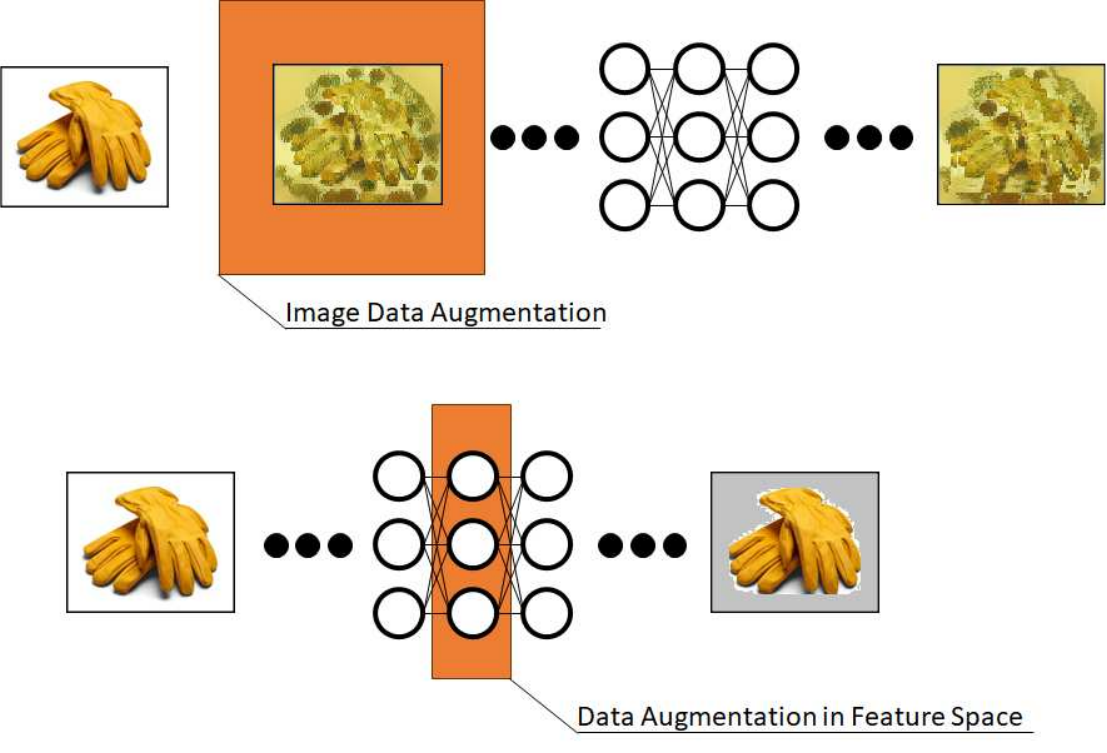}
        \caption{Feature data augmentation techniques extract the image's feature representation and apply the transformation in the feature space}
        \label{img:data-augmentation-technique-feature-augment}
    \end{minipage}\hfill
    \begin{minipage}{0.50\textwidth}
        \centering
        \includegraphics[width=0.95\textwidth]{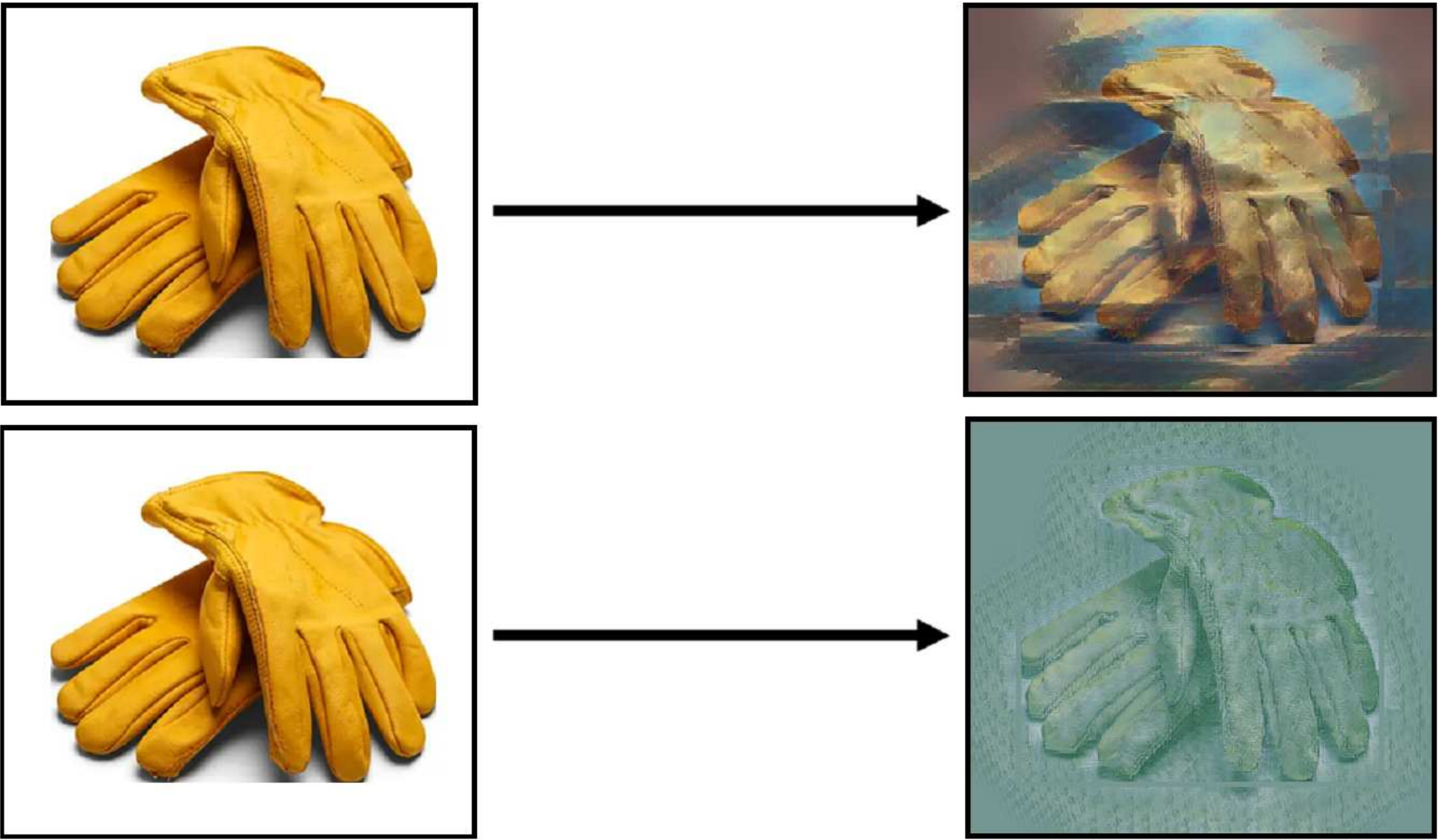}
        \caption{Neural style transfer methods generate synthetic images by modulating image characteristics while preserving the original semantic information}
        \label{img:data-augmentation-technique-style-transfer}
    \end{minipage}
\end{figure}

\paragraph{Feature Space Augmentations}

While traditional DA techniques have focused on applying transformations on the raw input data, feature augmentation techniques have also been used to enhance the learning capabilities of a model by manipulating the extracted feature representations. This can involve methods such as introducing new features, transforming existing ones, or modifying specific feature values. For instance, the work in \cite{wang2023feature} proposes the Information Fusion Rectification method to identify relevant base class prototype features from the query set using cosine similarity. The method leverages the relationship between base, support, and query samples and rectifies the support set's distribution using cosine similarity metrics as weights, generating new samples on the extracted feature space relevant to the provided task. Figure \ref{img:data-augmentation-technique-feature-augment} shows an example of applying feature data transformations.

\paragraph{Neural Style Transfer}

A different way to perform DA could be by leveraging modern techniques such as Neural Style Transfer, which offers a unique approach focusing on style-based variations. In \cite{jackson2019style}, the use of random style transfer is proposed by introducing randomness to texture, contrast, and color while preserving the underlying shape and semantic content of data. This is achieved by adapting an arbitrary style transfer network to perform style randomization, with target style embeddings being sampled from a multivariate normal distribution instead of being computed from a specific style image. Figure \ref{img:data-augmentation-technique-style-transfer} illustrates an example of generating synthetic images following different styles. Finally, diffusion models have also been utilized in this context. Specifically, in \cite{feng2023diverse}, a pretrained CLIP encoder is used to extract latent features fed to Stable Diffusion to generate synthetic images. Data diversity is ensured by filtering generated images based on cosine similarity, while augmentation is enabled in image and feature spaces.

Overall, DA methods are an effective way of dealing with FSL due to their simplicity, flexibility, and facile combination with approaches such as meta-learning. However, most of these methods leverage domain knowledge and focus on image data, thus lacking transferability between different domains and problems. Consequently, more research towards developing domain-agnostic DA approaches is necessary.

\subsection{In-Context Learning} \label{in-context-learning}

Although not explicitly trained for it, it has been demonstrated that LLMs are capable of generalizing to novel tasks when given just a few demonstrations. This emergent few-shot ability of LLMs was first examined in \cite{brown2020language}, where scaling the model parameters and the number of training data leads to the creation of models capable of adapting to novel tasks without further parameter finetuning. Ever since, similar FSL capabilities have also been demonstrated under various model settings, such as bidirectional LLMs \citep{patel2022bidirectional} and different tasks, including machine translation \citep{garcia2023unreasonable}. In the literature, this few-shot ability has been referred to as in-context learning \citep{dong2022survey, zhou2024mystery} (ICL) since the model leverages a few examples that form a demonstration context to generalize within it. In-context learning can be seen as a learning mechanism based on analogies where the model draws on its prior knowledge and forms analogies between the provided demonstrations and the query at hand.

Comparing ICL and standard FSL, one of the major differences between the two approaches is their effect on the model parameters. More precisely, the main advantage of ICL is that the model is able to adapt to a new task without having to finetune its parameters through gradient descent. Simply by structuring a prompt that provides the context for the task through input-output pairs of examples, the model can generalize to the task, while in FSL, the model's parameters need further adaptation. As a result, ICL significantly reduces computing power and time associated with novel task adaptation. For instance, based on these few-shot capabilities, an incremental stage-wise training of language models has been proposed that enables faster and more efficient training and adaptation \citep{reddi2023efficient}. Moreover, another difference is that FSL models are explicitly trained to learn how to rapidly adapt to new tasks with just a few training samples, while ICL naturally emerges in LLMs without the need for explicit optimization. Finally, an additional advantage of ICL is its interpretability since the demonstrations are provided in natural language, thus facilitating the incorporation of human knowledge in them.

Undoubtedly, these desirable characteristics of ICL have rendered LLMs particularly attractive in various applications where data is limited. In \cite{meng2023tuning}, an autoregressive LLM is finetuned through prefix-tuning to generate new data. Consequently, the generated data is used to augment the original dataset in an attempt to render supervised learning feasible. In \cite{dai2022promptagator}, LLMs are used as few-shot query generators to enhance information retrieval in cases where training data is absent. Notably, the proposed method manages to outperform models pretrained on large question-answering datasets by simply using eight examples as demonstration input for the LLM data generator. Furthermore, LLMs have been leveraged as few-shot abductive reasoners \citep{shi2023language}. In that case, given a few annotated examples, their goal is to suggest possible causes for an event as part of an event sequence modeling pipeline. Finally, in recent years, ICL has also been applied to various modalities beyond text, using vision \citep{foster2023flexible} and language-vision \citep{tsimpoukelli2021multimodal} models, demonstrating its potential as a new general form of FSL.

\subsubsection{Theoretical Understanding of In-Context Learning}

In recent years, there have been several attempts to shed light on the mechanisms that enable ICL in LLMs. On a theoretical level, various studies have analyzed how specific components within LLMs contribute to the emergence of ICL. Specifically, in \cite{olsson2022context}, the authors explain how certain attention heads in the model's architecture, called induction heads, are responsible for ICL by reproducing previously occurring token sequences for next-token prediction. More recently, this direction has been further explored using a Markov chain modeling task \citep{edelman2024the}. At the same time, another recent study has examined ICL through the lens of function vectors, which are responsible for transmitting information regarding the demonstration examples across the transformer modules \citep{toddfunction}.

Given the similarities between ICL and meta-learning for FSL, in terms of their ability to adapt to novel tasks using a minimal number of samples, various studies have attempted to frame ICL as a type of gradient-based meta-learning, where the forward pass of the transformer modules in ICL approximates performing finetuning within each task's support set in FSL \citep{von2023transformers}. Specifically, in \cite{dai2023can}, GPT-based models are viewed as meta-optimizers and ICL as implicit model finetuning based on a recasting of transformer attention as a meta-optimization module. However, recent studies have suggested that these similarities may not be sufficient to fully explain ICL due to differences between the inner workings of ICL and gradient descent \citep{deutch2024context}, necessitating further study in that area.

Another line of theoretical research on ICL has focused on analyzing it as a form of function regression learning, where the model learns how to fit a regression function to the provided demonstration examples. In that direction, most existing works have focused on learning linear regression functions from the given examples \citep{garg2022can, akyureklearning}. At the same time, a more recent study has also extended this framework to a broader setting by introducing a compositional structure that combines learning a fixed representation function with a linear regression function \citep{guo2024how}.

Lastly, in \cite{xie2021explanation}, ICL is presented as a form of implicit Bayesian inference, and the effect of the training data distribution is underlined. The same framing has also been combined with the regression function learning \citep{panwar2024incontext} and gradient-based meta-learning \citep{jeon2024an} frameworks in an attempt to provide a broader and unified perspective of how ICL works.

\subsubsection{Empirical Insights on In-Context Learning}

Parallel to the works providing a theoretical framework for ICL, various studies have focused on identifying and quantifying the effect of different influencing factors from an empirical perspective. More specifically, model size \citep{kaplan2020scaling} has been pinpointed as an influencing factor, yet it is inadequate to illustrate the intrinsic properties that lead to the emergence of ICL. Similarly, other model properties dubbed as possible influencing factors have been computation \citep{wei2022emergent}, and the objectives selected during model pretraining \citep{tay2023ul}. In terms of the data used during model pretraining, attributes such as their domain \citep{shin2022effect}, distribution \citep{chan2022data}, and diversity \citep{raventos2024pretraining} have been examined, underscoring the importance of the pretraining data for the emergence of ICL.

Since LLMs are not explicitly trained to perform ICL, this emergent property can also be influenced by various factors during inference. Specifically, attributes of the examples used as demonstrations in the ICL setting, such as their diversity \citep{an2023context}, their similarity with the test query \citep{liu2022makes}, and their ordering \citep{liu2024lost} have been shown to impact ICL performance. Moreover, another critical factor during inference affecting ICL is the accuracy of the mapping between inputs and labels in the demonstrations and the samples used during pretraining \citep{kossen2023context}.

However, there are cases where findings seem to be contradictory regarding the actual impact of the various examined factors. For instance, while in \cite{kandpal2023large} a correlation between the relevance of the pretraining data and the examined downstream task is identified, in \cite{han2023understanding}, the authors show that this factor does not influence the effectiveness of ICL. Additionally, although initial findings have demonstrated that the input-label mapping does not influence ICL \citep{min2022rethinking}, subsequent studies have revealed that a consistent input-label mapping between pretraining data and the demonstrations used in ICL can lead to improved performance \citep{kossen2023context}.

\subsubsection{Boosting In-Context Learning with Meta-Learning}

An interesting aspect of ICL is that despite being itself a form of FSL, it has recently been combined with other FSL paradigms and, most notably, meta-learning to enhance its performance. Specifically, ICL is related to prompt learning, which allows for the use of learnable prompts instead of hand-crafted ones. In this context, meta-learning can be utilized to efficiently learn better prompts across various downstream tasks, thus enhancing the performance of ICL. Although LLMs already demonstrate implicit few-shot capabilities, their performance in FSL scenarios can be further increased through explicit training on few-shot formulated tasks. For instance, in \cite{min2021metaicl}, a meta-training scheme based on formulating few-shot tasks is proposed to enhance the ICL capabilities of LLMs. Additionally, in \cite{hou2022metaprompting}, an approach adapting MAML for prompt tuning is proposed, which meta-learns an effective prompt initialization that enables fast adaptation to multiple downstream tasks. Lastly, a novel approach has recently been proposed aiming to improve ICL through meta-ICL \citep{coda2023meta}. More specifically, although ICL allows for rapid adaptation to downstream tasks using a minimal number of demonstrations, different tasks may benefit from different priors and ICL strategies. By introducing an additional level where ICL is performed across sets of downstream tasks, the model is able to adapt its priors and learning strategies and perform ICL more effectively within each group of tasks.

\subsubsection{Limitations}

Although the benefits of ICL have been both theoretically explored and practically demonstrated, there are still various limitations pertinent to its inherent nature. As discussed earlier, ICL can be unstable, influenced by factors such as the prompting format, the examples provided, and even their ordering \citep{zhao2021calibrate}. To address these issues, different calibration methods have been proposed to make LLMs less sensitive to these parameters. For instance, in \cite{zhao2021calibrate}, various types of biases in prompts are identified, and a contextual calibration approach is proposed based on content-free test inputs. In \cite{han2022prototypical}, prototypical networks are utilized to calibrate the decision boundary of LLMs for classification in the ICL setting. This is achieved by creating prototypical clusters based on the support set samples using a Gaussian Mixture Model. Consequently, when combined with this type of calibration, LLMs demonstrate increased robustness regarding prompt ordering, class imbalance, and size of estimate set. Lastly, a recent study has posited that ICL can be viewed as a byproduct of using nonlinear or discontinuous metrics rather than an emergent property of LLMs \citep{schaeffer2024emergent}. Overall, this lack of clear consensus regarding a comprehensive theoretical and empirical justification of ICL highlights the need for further research to shed light on this novel and distinct form of FSL.

\section{Few-Shot Learning Beyond Supervised Learning Settings} \label{beyond-supervised-FSL}

Something that can be easily inferred from the models presented so far is that almost all of these methods have been developed for use in a supervised setting. However, this greatly limits the applications in which FSL can be used since many ML applications rely on unlabeled data. As a result, various novel approaches have emerged in recent years that aim to address this limitation by extending the FSL paradigm to different types of problems. At the same time, the close relationship of FSL models with techniques developed within other learning paradigms, such as self-supervised learning or multi-task learning, has led to the bridging of these approaches and the development of new hybrid methods that combine the best of both worlds.

\subsection{Semi-Supervised FSL}

\begin{figure}[h]
 \begin{center}
  \includegraphics[width=0.99\textwidth]{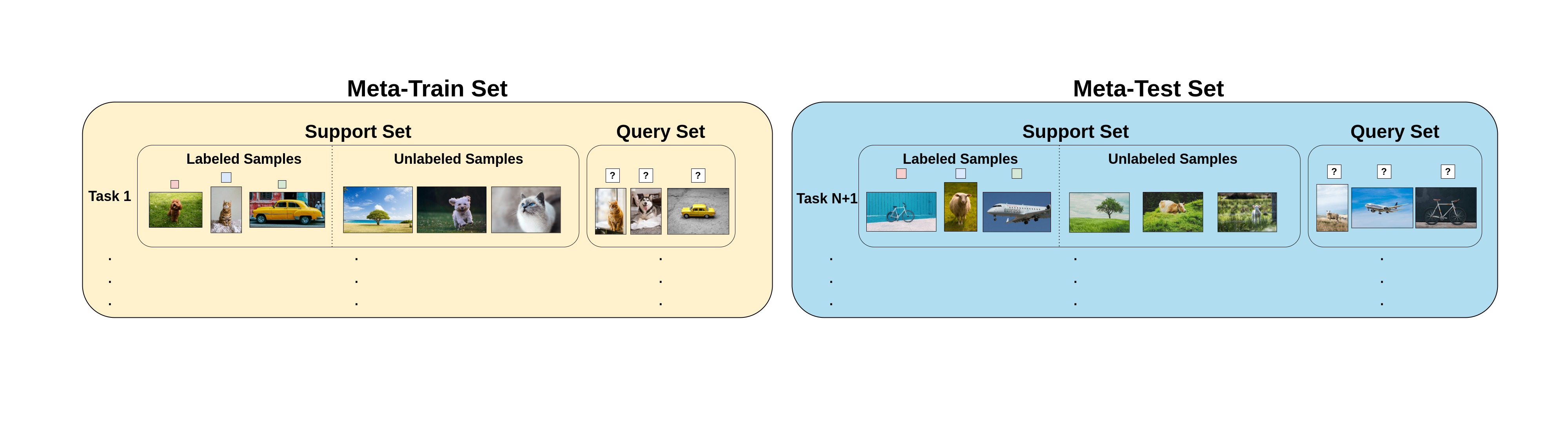}
 \end{center}
 \caption{Overview of Semi-Supervised FSL, where each task's support set includes both labeled and unlabeled samples}
 \label{img:semi_supervised_fsl}
\end{figure}

The FSL problem setting presented in Section \ref{formulation} assumes that all samples in each task's support set are labeled. Specifically, for any task $T$ in the \textit{meta-train set} and \textit{meta-test set} the standard FSL setting assumes that a task's $T$ support set $D_T^S$ consists of labeled samples, i.e., $D_T^S = \{ (x_i, y_i) \}_{i=0}^{N_{T_L}}$, where $N_{T_L}$ is the total number of labeled samples in the support set. In contrast, semi-supervised FSL addresses a more generalized setting that is more eminent under real-world conditions, where the support set includes both labeled samples $D_l=\{(x_{l,i}, y_{l,i})\}_{i=1}^{N_l}$ and unlabeled samples $D_u=\{x_{u,i}\}_{i=1}^{N_u}$. In this context, the support set can be expressed as $D_T^S=D_l \cup D_u$. It is also worth noting that $N_l$, which is the number of labeled samples in the support set, is typically small, while $N_u$, which is the number of unlabeled samples in the support set, is significantly larger. Figure \ref{img:semi_supervised_fsl} provides an overview of the above problem formulation. Overall, the objective of semi-supervised FSL is to improve performance for downstream FSL tasks by leveraging both the labeled and unlabeled data of each task.

One of the first works to incorporate unlabeled data within each task's support set was in \cite{ren2018meta}, where the authors propose extending prototypical networks for semi-supervised FSL. Specifically, soft k-means clustering is used, where the initial clusters are the prototypes derived by using the labeled samples, and the refined prototypes are obtained by incorporating the clustered unlabeled samples. The framework is also extended to include unlabeled samples belonging to distractor classes by introducing an additional cluster to deal with distractor samples and proposing a soft masking mechanism in the case of multiple distractor classes. 

In general, similar to supervised learning, directly adapting approaches commonly used in standard semi-supervised learning formulations to the FSL setting would yield inferior results due to the limited availability of labeled training samples within each task. In that setting, a common approach to effectively incorporate the available unlabeled data in each task is by propagating labels from the available labeled support set data to the unlabeled data. Specifically, some common approaches have focused on embedding both labeled and unlabeled samples into a graph and leveraging the manifold structure of the data to assign pseudo-labels to the unlabeled samples \citep{lazarou2021iterative, liulearning}, thus increasing the number of labeled samples available for adaptation within each task.

However, predicting pseudo-labels for unlabeled samples can be challenging, especially when only a limited amount of labeled data is available. To tackle this issue, the authors in \cite{wei2022embarrassingly} proposed reversing the problem of pseudo-labeling prediction to predicting the negative pseudo-labels for each unlabeled sample based on the class that is less probable to belong to. The method is applied iteratively to all unlabeled samples, and the actual positive pseudo-labels are eventually retrieved by eliminating all assigned negative pseudo-labels for each sample.

\subsection{Unsupervised FSL}

\begin{figure}[h]
 \begin{center}
  \includegraphics[width=0.99\textwidth]{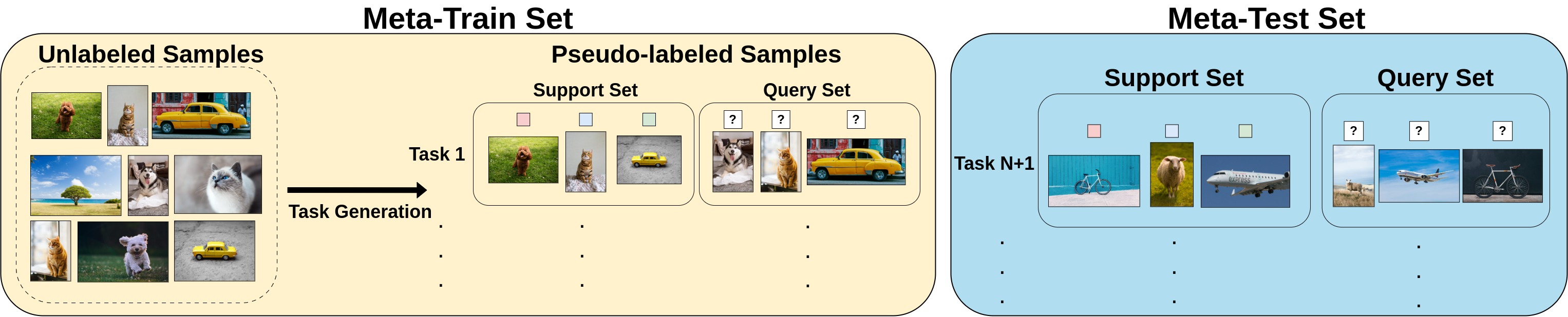}
 \end{center}
 \caption{In the Unsupervised FSL setting, meta-train set tasks are generated using pseudo-labels assigned to the unlabeled samples available during meta-training}
 \label{img:unsupervised_fsl}
\end{figure}

A major principle underlying all FSL approaches, regardless of their specific formulation, is their ability to leverage prior knowledge gained from different tasks or data sources to adapt to novel tasks quickly using only a small number of training samples. In this context, unsupervised FSL can be seen as a natural extension of supervised and semi-supervised FSL, where a large amount of data is available for meta-training, yet none of it is labeled. More specifically, in unsupervised FSL we consider that the $meta-train$ set contains only unlabeled samples, $D_{meta-train}=\{x_i\}_{i=1}^{N_{total}}$, where $N_{total}$ is the total number of unlabeled samples. Since the classes of the tasks in the $meta-test$ set are unknown during meta-training, it is impossible to propagate pseudo-labels from the labeled samples in the $meta-test$ set to the unlabeled samples. Consequently, techniques from semi-supervised learning cannot be successfully applied in this setting, necessitating a different approach to extracting prior knowledge from the unlabeled samples. Figure \ref{img:unsupervised_fsl} provides a high-level overview of the task-generating procedure during meta-training in the Unsupervised FSL problem formulation.

For this specified setting, most approaches combine unsupervised learning and meta-learning to enable rapid learning within different downstream tasks with a few labeled samples. These methods generally utilize information extracted from the unlabeled data and learn how to leverage the underlying shared structure across different tasks. A common approach to dealing with unsupervised FSL is creating synthetic tasks with unlabeled data to enable meta-learning on them. CACTUS \citep{hsu2018unsupervised}, one of the first methods in that direction, performs multiple executions of k-means in the feature representation space to obtain partitions that can be used as pseudo-classes, and uses these partitions to construct tasks used in training supervised meta-learning models. In \cite{khodadadeh2020unsupervised}, a generic, domain-independent approach based on generative models is proposed for task generation. These models are trained to model the unlabeled dataset's latent space, and same-class samples are created by sampling in the proximity of a point in the latent space. Similar approaches for unsupervised FSL have also been applied to tabular data \citep{nam2023stunt}. The proposed method, STUNT, leverages a large unlabeled dataset to create diverse tasks with pseudo-labels using k-means clustering and feature perturbations. Inference is then performed on a small labeled dataset by a Prototypical Network trained on the generated tasks.

Finally, apart from task generation, other works have also focused on directly learning the multi-modality within each randomly sampled task. In \cite{kong2021unsupervised}, a top-down generative model that learns unknown labels through categorical latent variables is proposed, that consists of an Energy-Based Model (EBL) that couples latent representations with class labels. Interestingly, the label can be inferred from the latent representations, allowing the model to learn the structure of different classes for different tasks, even without access to class labels.

\subsection{Self-Supervised FSL} \label{self_supervised_fsl}

Although meta-learning and self-supervised learning constitute different learning paradigms and are generally used for different problems, they share a common underlying goal: to extract knowledge from upstream tasks and apply it to learning downstream tasks more efficiently. Additionally, similar to the methods for unsupervised FSL, self-supervised learning utilizes unlabeled data during training and uses them to create pretext tasks to train the models and improve their performance in downstream tasks with limited labeled samples. Consequently, various approaches leverage that close relationship \citep{ni2021close} between these learning paradigms to improve FSL, particularly in the unsupervised setting.

\begin{figure}
    \centering
    \begin{minipage}{0.59\textwidth}
        \centering
        \includegraphics[width=0.92\textwidth]{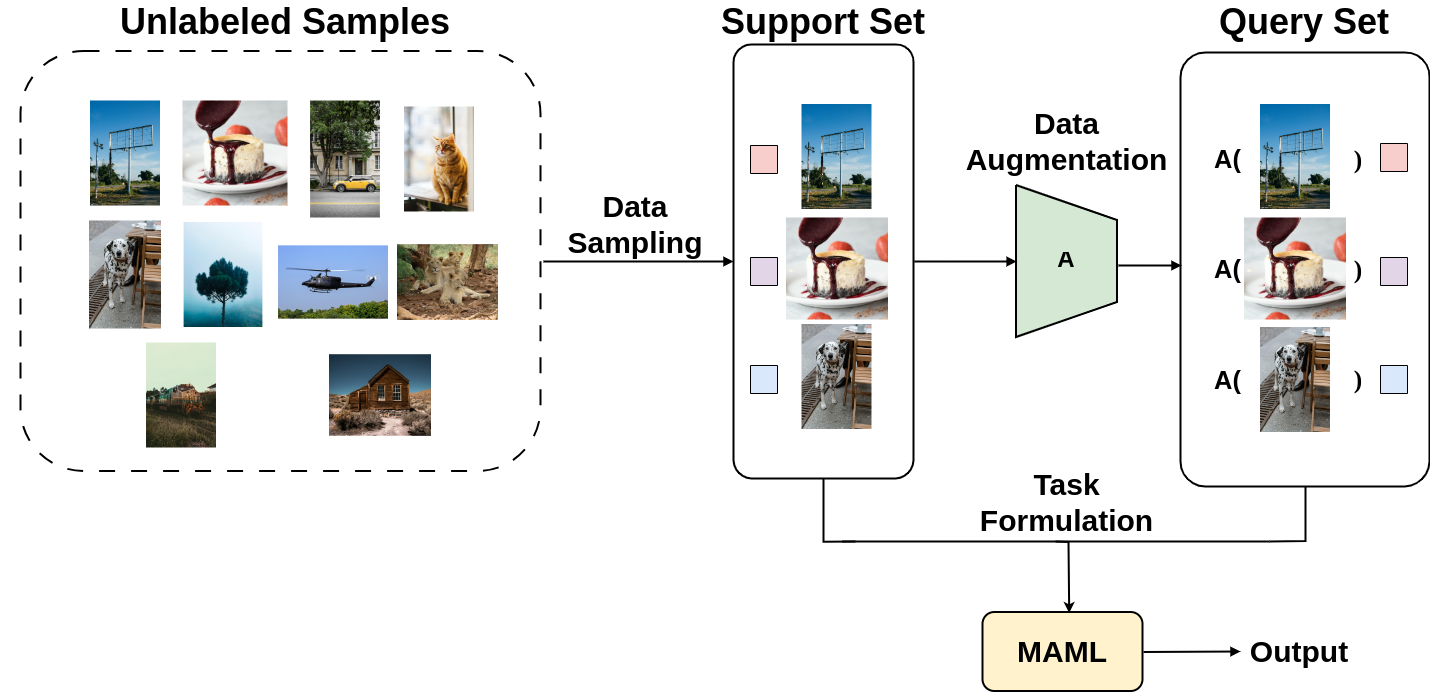}
        \caption{In the UMTRA architecture \citep{khodadadeh2019unsupervised}, one-shot tasks are generated by matching support set samples with their augmented counterparts in the query set}
        \label{img:hybrid-self-supervised}
    \end{minipage}
    \begin{minipage}{0.38\textwidth}
        \centering
        \includegraphics[width=0.92\textwidth]{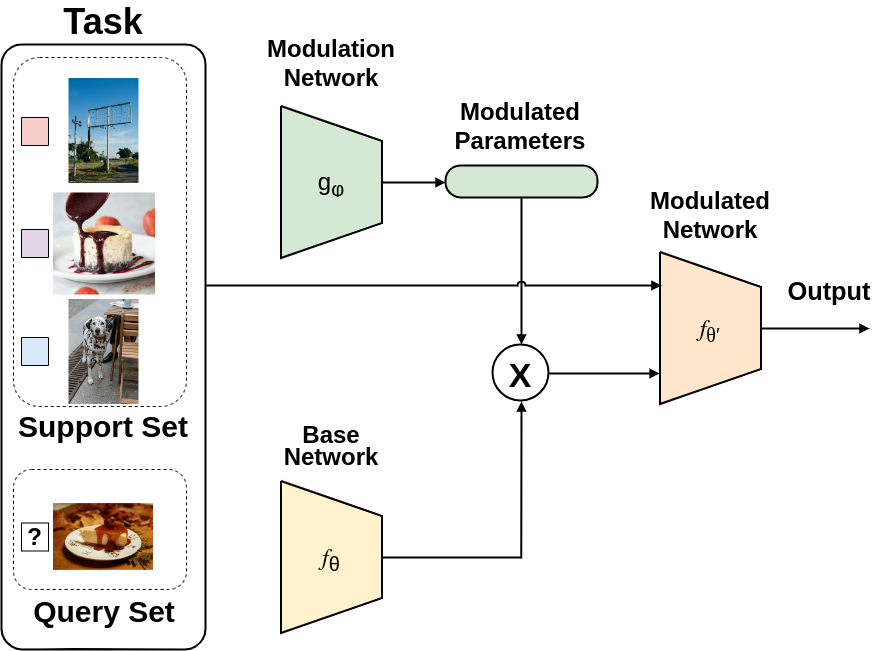}
        \caption{In the KML architecture \citep{abdollahzadeh2021revisit}, the model parameters are modulated based on the concept of transference from multi-task learning}
        \label{img:hybrid-multitask}
    \end{minipage}
\end{figure}

In general, many of these techniques use an unlabeled dataset to formulate few-shot tasks that are used to train a model that performs FSL in the labeled dataset. For instance, in \cite{khodadadeh2019unsupervised}, one-shot tasks are generated by assuming that each selected sample belongs to a different class, and the corresponding query set is created by applying basic image DA techniques. Figure \ref{img:hybrid-self-supervised} presents a high-level overview of the model's architecture. Additionally, in \cite{jang2023unsupervised}, each unlabeled sample is considered a query, and the top-k similar samples are selected to be in the support set using a contrastive loss. On the other hand, in \cite{lee2023self}, task generation from unlabeled samples has been approached through set representation learning. Specifically, a contrastive loss that considers instance-level, set-level, and cross-similarity between positive pairs and instances is used during meta-training. At test time, a representation is learned for each set, which is then used as initial weights for a linear classifier that is fine-tuned on each task's support set.

Lastly, contrastive learning has been employed in the context of few-shot classification of partial views of images \citep{jelley2023contrastive}. In that case, the partial image views are used as samples within each task, and each image constitutes a different class. Finally, a generalized version of Prototypical Networks that uses a product of experts and leverages contrastive learning is used for classification.

\subsection{Multi-Task FSL}

Another combination of approaches in FSL includes that of multi-task learning (MTL) and meta-learning. In general, meta-learning can be seen as a more generalized version of MTL, aiming to adapt to previously unknown tasks. In MTL, a specific set of tasks is defined during training, and the model is trained to generalize to these tasks simultaneously. In contrast, in standard FSL and meta-learning settings, the tasks seen during meta-training and meta-testing differ, and the number of tasks in meta-testing may also vary.

Yet, despite these differences, MTL can provide valuable elements and techniques that meta-learning can benefit from. For instance, in \cite{abdollahzadeh2021revisit}, the idea of transference from MTL is extended to quantify the transference of knowledge from training tasks to test tasks in meta-learning. Additionally, a modulation mechanism (KML) based on hard parameter sharing that extends that of Multimodal MAML \citep{vuorio2019multimodal} to mitigate negative transfer between tasks is proposed. A high-level overview of KML's architecture can be seen in Figure \ref{img:hybrid-multitask}.

Moreover, in \cite{ye2021multi}, MAML's bi-level optimization formulation is extended by introducing multiple objective functions at the upper level, and the Multiple Gradient Descent Algorithm (MGDA) is used to tackle the existence of multiple objectives. Finally, in \cite{yu2022enhancing}, meta-learning is cast as a multi-objective optimization problem, where each task's loss is handled as a different objective. Notably, this method can help mitigate the compromising phenomenon, where the solution focuses on minimizing the losses of only a small subset of tasks and converges faster than MAML.

\subsection{Cross-Domain FSL}

\begin{figure}[h]
 \begin{center}
  \includegraphics[width=0.99\textwidth]{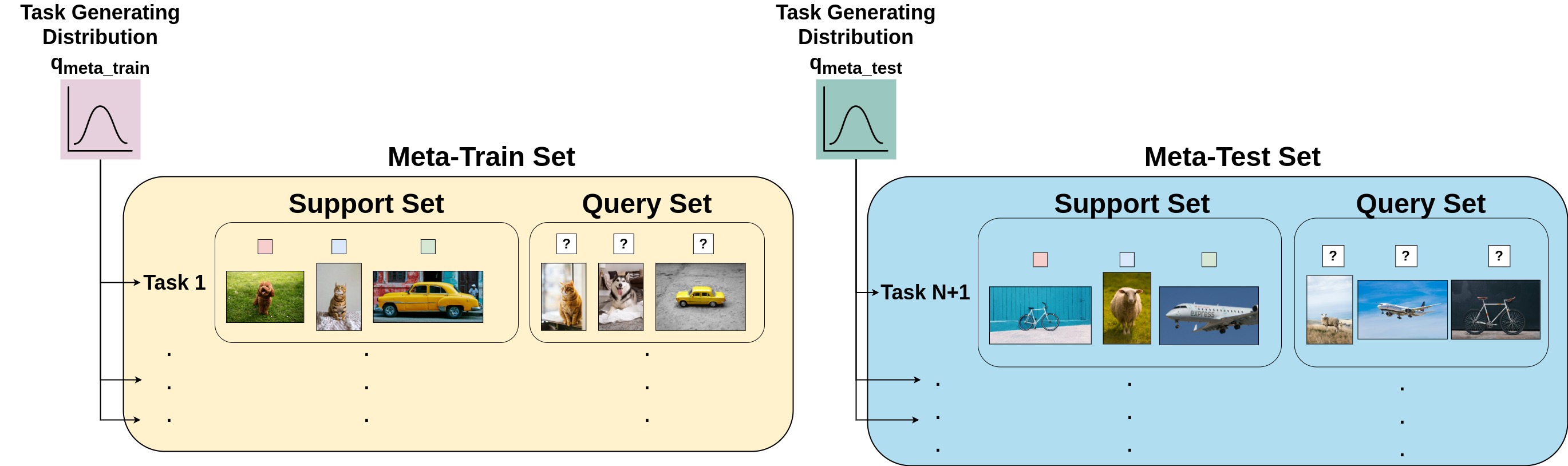}
 \end{center}
 \caption{Overview of Cross-Domain FSL, where the meta-train set tasks and meta-test set tasks are generated from different task-generating distributions}
 \label{img:cross_domain_fsl}
\end{figure}

Cross-Domain FSL (CD-FSL), which combines FSL's knowledge transfer across tasks with different classes with Domain Adaptation's knowledge transfer across different data distributions, has emerged as a new learning paradigm in recent years. More specifically, CD-FSL aims to leverage knowledge extracted from the base tasks to adapt to new tasks that have both novel classes and also follow a different distribution \citep{wu2023domain}.

In the problem formulation of FSL in Section \ref{formulation}, an assumption was made that all tasks in both \textit{meta-train} and \textit{meta-test} sets are derived from the same task distribution $q(T)$. However, in real-world applications, it is possible for the tasks in the \textit{meta-test} set to originate from a different task distribution compared to those in the \textit{meta-train} set. Consequently, in CD-FSL, the tasks that belong to the \textit{meta-train} set $D_{meta-train}=\{D_i\}_{i=1}^N$, where $D_i$ is the dataset corresponding to a task $T_i$ are drawn from a task distribution $q_{meta-train}(T)$, i.e., $T_i \sim q_{meta-train}(T)$, while the tasks that belong to the \textit{meta-test} set $D_{meta-test}=\{D_j\}_{j=1}^M$ are drawn from a task distribution $q_{meta-test}(T)$, i.e., $T_j \sim q_{meta-test}(T)$, and $q_{meta-train} \neq q_{meta-test}$. Figure \ref{img:cross_domain_fsl} illustrates more clearly how tasks in $D_{meta-train}$ and $D_{meta-test}$ originate from different distributions. Although more challenging, this setting is particularly relevant since, in practice, it is typical for the base and target datasets of FSL to originate from different domains \citep{zheng2023cross}. For instance, a model might be trained on a large dataset of natural images, such as ImageNet, and be adapted and evaluated on a smaller dataset that is fundamentally different, such as medical or satellite images. 

A common challenge in CD-FSL is that representations on the target dataset are often biased due to the data volume discrepancy, making it difficult to obtain robust models across different domains. To address this issue, recent approaches have focused on mitigating cross-domain bias during both the development and evaluation stages. At the development level, CD-FSL includes models that are explicitly trained to handle few-shot domain adaptation/generalization, with methods such as finetuning a model's projection and prediction heads \citep{jang2023unsupervised}, retraining the feature extractor using an inverted Firth Bias regularizer \citep{wang2022revisit}, and using knowledge distillation \citep{zheng2023cross}. The evaluation level includes few-shot cross-domain adaptation of models, even if they were not explicitly trained for this task, which is useful in a wide variety of applications such as visual question answering, temporal point processes, neural architecture search, and drug discovery \citep{schimunek2023context}.

Overall, robustness against domain shifts can be a potential step towards achieving Broad AI \citep{schimunek2023context}. At the same time, cross-domain evaluation in meta-learning could help measure overfitting at the meta-level by attributing generalization to learning or memorization at the task level.

\subsection{Few-Shot Federated Learning}

\begin{figure}[h]
 \begin{center}
  \includegraphics[width=0.99\textwidth]{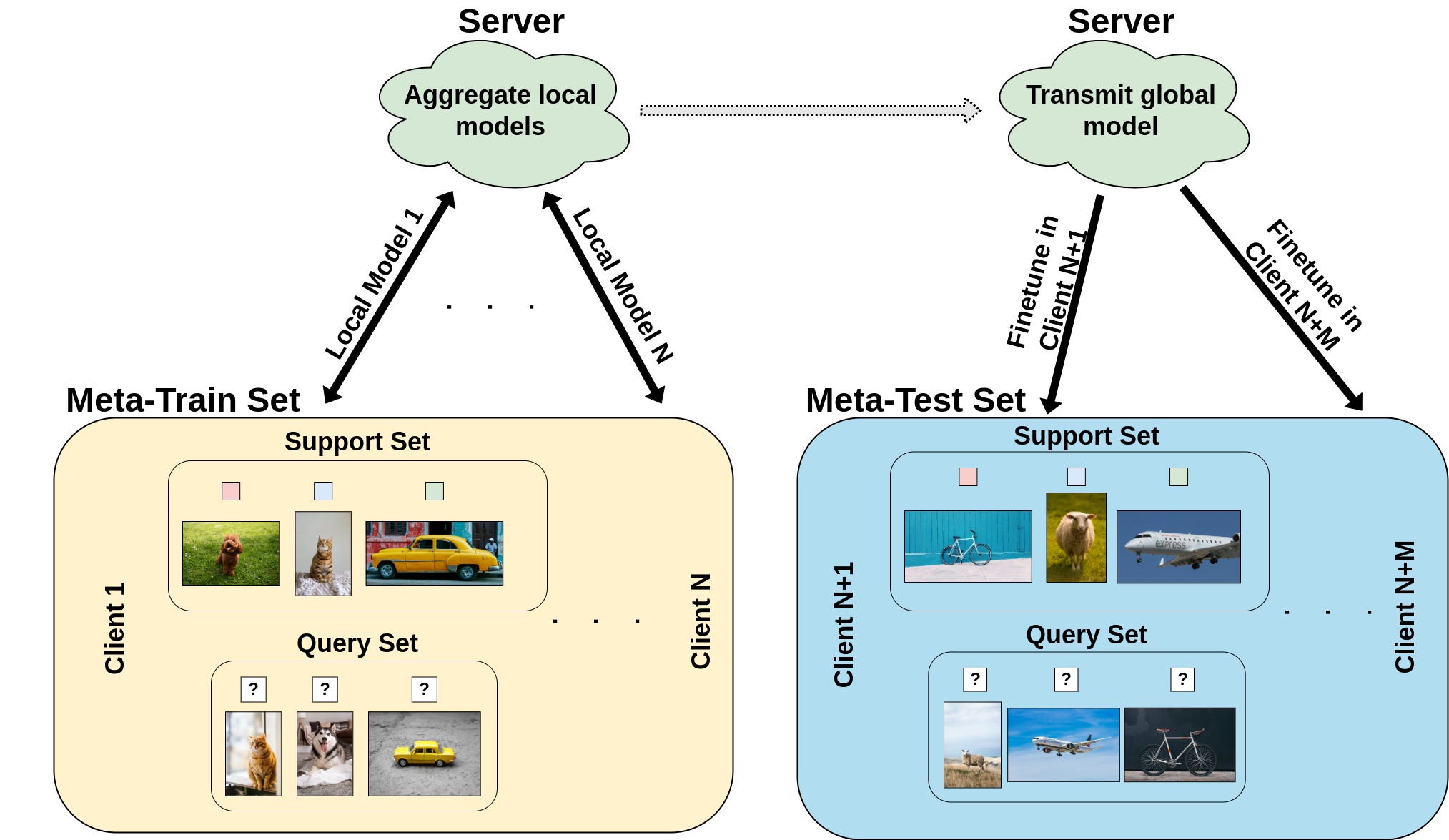}
 \end{center}
 \caption{Overview of Few-Shot Federated Learning, where the standard FSL framework is extended to consider each client as a separate task}
 \label{img:federated_fsl}
\end{figure}

In recent years, decentralized types of learning like Federated Learning \citep{ayeelyan2025federated} (FL) have been extensively researched to leverage scattered data across multiple machines while preserving privacy. In the standard FL setup, there are multiple clients, each having its own private dataset, denoted as $D_i$ for the $i$-th client. Under this setup, the objective is to learn a global model that performs well across all clients' datasets. To achieve this, various approaches have been developed that aim to aggregate  each client's model weights to create a global model that generalizes to all clients' data simultaneously. This way, FL reduces communication costs and enhances data privacy by sharing only model parameters between servers and clients.

Given that there is sufficient data within each client's dataset, it has been demonstrated that these methods can yield strong results. However, FL's performance significantly declines in scenarios where data is limited for each client. To this end, Few-Shot Federated Learning (FS-FL) has been explored as a potential way to address these issues arising in federated applications with limited data. In FS-FL, each client is treated as a separate task with its own associated dataset. Consequently, the problem can be recast as an FSL problem that follows the same formulation as the one outlined in Section \ref{formulation}, with the additional consideration that there can be no data exchange between different tasks. Under this formulation, the FL's goal of learning a model that performs well in all clients' data is also recast as learning a model that can rapidly adapt to each client's dataset with minimal training samples. Figure \ref{img:federated_fsl} provides an overview of the FS-FL's problem formulation. Based on this recasting of FS-FL as an FSL problem, several approaches combining FL with meta-learning have been proposed, such as adjusting popular meta-learning algorithms for FL \citep{chen2018federated}, recasting FedAvg as a meta-learning algorithm for personalized FL \citep{jiang2019improving} and separately executing meta-learning's optimization loops on the server and clients \citep{fallah2020personalized}.

In the last few years, a plethora of different FS-FL methods have been developed for applications such as facial expression recognition \citep{shome2021fedaffect}, load forecasting \citep{tang2023privacy}, 3D MRI medical image classification \citep{he2023dealing}, fault diagnosis in industrial settings \citep{chen2023industrial}, and NLP \citep{dong2022fewfedweight}. A common theme across all of these applications is the sensitive nature of the involved data that necessitates privacy preservation, as well as the fact that this data is typically scarce, heterogeneous, and dispersed in different machines. Consequently, FS-FL is an ideal solution for these types of problems.

\subsection{Continual FSL}

\begin{figure}[h]
 \begin{center}
  \includegraphics[width=0.99\textwidth]{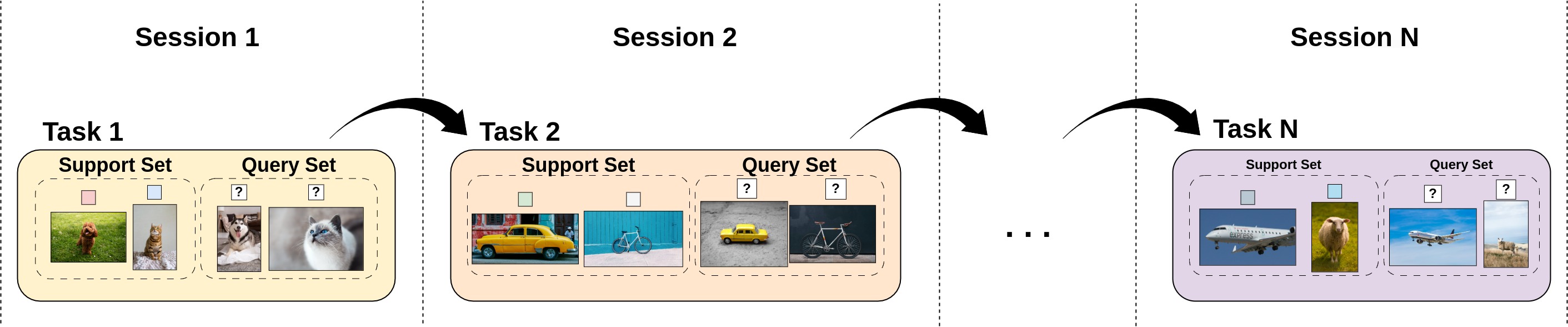}
 \end{center}
 \caption{In the Continual FSL framework, each task appears sequentially in a session-by-session manner}
 \label{img:continual_fsl}
\end{figure}

Following the recent trend of bridging the gap between FSL and related fields of study, FSL has also been applied in the context of Continual Learning \citep{de2021continual}. Continual Learning refers to the ability of a model to learn from an infinite stream of data, acquiring new knowledge along the way while also preserving previously gained information. In general, training in this setting is performed in a session-by-session manner, and one of the main challenges is overcoming catastrophic forgetting where the acquisition of new knowledge during subsequent sessions results in performance degradation on the knowledge accumulated during previous sessions. In the case of Continual FSL, each session consists of a single or a batch of different N-way-K-shot tasks with limited training samples. This scarcity of training data introduces an additional layer of difficulty since typical Continual Learning methods are susceptible to overfitting the new data. Figure \ref{img:continual_fsl} illustrates the session-by-session learning procedure that is typical in Continual FSL.

One of the first attempts to address Continual FSL recasts meta-learning in the online learning setting by developing a new algorithm called follow the meta-leader \citep{finn2019online} (FTML). FTML is the meta-learning version of the widely used follow-the-leader algorithm and extends MAML \citep{finn2017model} in the online learning setting. However, catastrophic forgetting is not explicitly addressed since all previous data is stored and is available in each session. Overfitting is dealt with by leveraging meta-learning, thus obviating the need to take many gradient steps to refine the model parameters within each task. Other notable approaches in Continual FSL include introducing a mixture of across-tasks shared parameters through a Dirichlet process mixture model to create a meta-learning model capable of incrementally learning heterogeneous tasks \citep{jerfel2019reconciling} and using Bayesian meta-learning with different meta-knowledge components encapsulating across-tasks knowledge to enable continual learning of sequentially available heterogeneous tasks \citep{wu2023adaptive}.

\subsection{Class-Incremental FSL}

\begin{figure}[h]
 \begin{center}
  \includegraphics[width=0.99\textwidth]{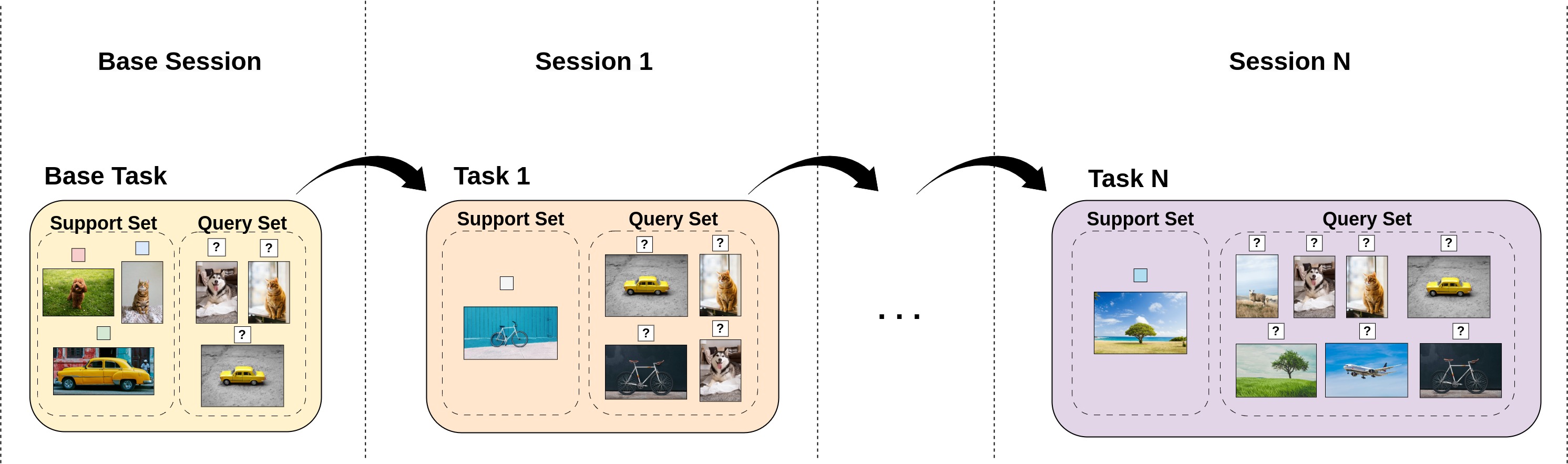}
 \end{center}
 \caption{Overview of Class-Incremental FSL where in each session additional classes are added, with only limited support set samples available, and the model is tasked with predicting from all classes seen up to that point}
 \label{img:class_incremental_fsl}
\end{figure}

Much of the recent research revolving around Continual FSL has mainly focused on Class Incremental FSL \citep{zhang2023few} (CIFSL) where in each session, a new set of classes is introduced, and the learning model should be able to infer from all classes encountered up to that point. This type of problem differs from the Continual FSL paradigm in that the first session (base session) typically consists of a task with a sufficiently large number of training samples. In contrast, the subsequent tasks in the incremental sessions are structured as N-way-K-shot tasks with only a limited number of samples. Figure \ref{img:class_incremental_fsl} illustrates the incremental nature of the Class-Incremental FSL's problem formulation.

In \cite{song2023learning}, a contrastive supervised loss is used to ensure adequate separation of the base classes and facilitate generalization to novel classes during the incremental sessions. Another common approach in class incremental learning is to leverage knowledge distillation to adapt to new sessions. However, standard knowledge distillation methods fail in the FSL setting as they tend to overfit novel classes with only a few samples. To mitigate this issue, a combination of two teachers has been proposed to generate the student network: one derived from the base session capturing base knowledge and one from the previous session capturing newly acquired knowledge \citep{zhao2023few}. Finally, various recently introduced methods for CIFSL draw inspiration from techniques and concepts in other areas of machine learning, such as recursive analytical learning \citep{zhuang2023gkeal}, neural collapse \citep{yang2023neural}, and the Lottery Ticket Hypothesis \citep{kang2022soft}.

\section{Few-Shot Learning Applications} \label{fsl_applications}

Even though academic research has traditionally focused on increasingly large-scale curated datasets for deep learning, collecting, storing, and processing such large volumes of data typically requires a large storing capacity and computational power, which is rarely available in real-world applications. At the same time, considerations such as high data collection costs, data privacy issues, and small datasets scattered across different edge devices have rendered FSL a suitable candidate for a plethora of real-world applications in various fields. In the rest of this section, we examine some of these applications, focusing on recent advancements and innovative approaches.

\subsection{Computer Vision}

Traditionally, most of the research pertinent to FSL has focused on computer vision problems and specifically image classification \citep{snell2017prototypical, finn2017model}, primarily due to the existence of appropriate FSL datasets for training and evaluation. However, in recent years, novel FSL approaches have been developed aiming to target different compute vision problems, including image segmentation \citep{gkanatsios2023analogy}, object detection \citep{demirel2023meta} and 3D reconstruction \citep{kwak2023geconerf}. Finally, the usage of FSL has also been extended to video inputs \citep{yan2023two}.

\subsection{Reinforcement Learning}

One of the first applications for many of the aforementioned meta-learning techniques has been meta-RL, which involves learning to quickly adapt across various similar environments by finding an optimal policy across tasks, given only a few episodes in each one of them. This is a particularly useful skill for agents acting in dynamic environments, largely resembling human-like learning, and has been used in the context of imitation learning \citep{rakelly2019efficient}, visual navigation \citep{Ni2023MetaDiffuserDM}, multi-armed bandits \citep{sung2017learning}, robot arm manipulation \citep{wang2023offline} and control \citep{Ni2023MetaDiffuserDM}. Novel applications of meta-RL also include online meta-RL \citep{khattar2022cmdp}, black-box optimization \citep{ma2023metabox}, long-term memory from visual observations \citep{beck2023recurrent} and implicit language learning \citep{liu2023simple}.

\subsection{Natural Language Processing}

Another field of research where FSL has drawn increasing attention in the last few years is that of NLP, with some standard applications including machine translation \citep{garcia2023unreasonable}, and text \citep{han2022prototypical} and sentiment \citep{yu2018diverse} classification. More recently, FSL has also been extended to novel applications and tasks such as prompt tuning \citep{jiang2023effective}, visual question answering 
\citep{najdenkoska2023meta}, and dense retrieval \citep{dai2022promptagator}. Finally, based on LLMs' ability to handle multiple downstream tasks when given only a few examples in the form of prompts (see Section \ref{in-context-learning}), various similar FSL techniques have also been proposed for applications such as natural language inference including multiple downstream tasks like causal reasoning, textual entailment, word sense disambiguation, and question answering \citep{murty2021dreca}.

\subsection{Audio}

In recent years, FSL has been successfully applied in the field of acoustic signal processing, with most applications focusing on audio classification \citep{wang2022hybrid} and its extensions in the continual \citep{wang2021few} and class-incremental \citep{li2023few} learning settings. Additionally, FSL has also been utilized for sound event detection \citep{wang2020few}, while a promising application is that of speaker \citep{li2023few} and sound recognition \citep{chou2019learning}, which can be effectively used in mobile devices and can benefit people with hearing difficulties. 

\subsection{Generative AI}

One of the most prominent and broadly expanding fields where FSL can be extremely useful is that of Generative AI. The ability to generate novel content given only a few reference samples has been extensively explored in the context of computer vision with recent applications, including image generation \citep{zheng2023my}, style transfer \citep{tang2023master}, and semantic image synthesis \citep{careil2023few}. FSL has also been utilized for generative purposes in different modalities such as time series \citep{jiang2022sequential}, video \citep{li2023one}, speech \citep{kumar2020few}, and music \citep{liang2020dawson} generation. Few-shot generation across different modalities has also gained much popularity recently via applications such as text-to-image \citep{zhou2022lafite2} and text-to-speech \citep{huang2022meta} generation.

\subsection{Automated Machine Learning}

Traditionally, Automated Machine Learning (AutoML) has been one of the most used testbeds for novel meta-learning techniques. AutoML aims to automatically find the best model architecture and hyperparameters for a given problem and dataset \citep{baratchi2024automated}. This reduces the need for domain-specific knowledge and extensive hyperparameter tuning, making machine learning more accessible and accelerating deployment. As for FSL in the context of AutoML, applications have mainly focused on Neural Architecture Search \citep{lee2023meta} (NAS), model selection \citep{park2022metagl}, and hyperparameter optimization \citep{maraval2023end}.

\subsection{Healthcare}

A natural field of application for FSL is when data collection can be difficult, costly, or even impossible. Consequently, healthcare is a perfect application field for FSL since most patient data is subject to privacy restrictions, yet high performance needs to be maintained under these critical settings. Recently, FSL has been successfully applied in medical image classification \citep{sun2023metamodulation}, medical image segmentation \citep{huang2023rethinking}, and 3D medical image segmentation in a federated setting \citep{he2023dealing}. FSL has also been critical in cases where more data acquisition is impossible, such as in rare disease classification \citep{sun2023metamodulation}, and drug discovery \citep{chen2022meta}. Finally, some novel approaches have focused on developing models that can handle multiple downstream medical tasks such as operational outcomes, anticipating lab test values, assignment of new diagnoses, and anticipating chest X-ray findings \citep{wornow2023ehrshot}.

\subsection{Recommendation}

A field that has particularly benefited from the use of FSL techniques is that of recommendation systems, by helping overcome major challenges, such as the cold-start problem, and improve model performance \citep{yu2021personalized}. Additionally, personalization is another area that has greatly benefited from FSL \citep{shysheya2022fit}. Specifically, by treating each user's data as a separate task, FSL can efficiently learn across all of them, even with limited data. Consequently, FSL has successfully been applied for personalized video retrieval \citep{yeh2023meta}, personalized recommendation \citep{wu2022meta}, and personalization in federated learning settings \citep{zhao2022personalized}.

\section{Recent Trends and Future Directions in Few-Shot Learning} \label{trends-challeges}

The rapid growth of FSL methodologies and applications has established it as a distinct area within the machine learning community. However, FSL is not an isolated field; various methodologies developed within FSL overlap with different areas of machine learning research. Consequently, FSL has been influenced by advancements and new challenges from both its own field and related research areas. For instance, the emergence of large foundation models has led to their application and evaluation in settings with limited available data. Additionally, more general research directions, such as developing models that exhibit traits of human-like intelligence and advancing AI to be more environmentally friendly, have also penetrated the FSL field in recent years. In this section, we provide a discussion of various trends that have been rising in recent years in the FSL field, as well as the existing and novel challenges that may shape future research directions in FSL.

\subsection{Few-Shot Evaluation of Foundation Models}

In recent years, the emergence of LLMs and foundation models, primarily in the visual and language domain, has created new opportunities for developing large-scale models capable of handling multiple downstream tasks with little or no adaptation. ICL of LLMs has been critical in that direction, and enabling similar capabilities on vision models via learning more powerful representations is the next logical step.

Based on these observations and the recent bloom of foundation models, one recent trend has been the evaluation and adaptation of these models regarding their FSL capabilities \citep{liu2024few}. Few-shot and even zero-shot performance is increasingly assessed both in the language \citep{brown2020language} and visual domain \citep{dosovitskiy2020image}, as well as in multimodal settings \citep{radford2021learning}. Additionally, variants of foundation models not explicitly developed for FSL \citep{kwak2023geconerf} and combinations of foundation models aiming to enhance few-shot performance \citep{zhang2023prompt} have also emerged.

Overall, testing the FSL capabilities of foundation models is a crucial step toward creating general models that can handle multiple tasks with minimal overhead. These foundation models should adapt quickly to novel tasks, alleviate overfitting, and work well in areas with limited data, such as in healthcare, leading to the democratization of AI and the mitigation of environmental damage due to extensive training.

\subsection{Meta-Learning in novel settings} 

Traditionally, most FSL applications have revolved around problems related to computer vision and RL. While FSL has also been applied to different domains, such as NLP, as seen in Section \ref{fsl_applications}, and under different problem formulations, such as unsupervised learning, as seen in Section \ref{beyond-supervised-FSL}, these constitute well-studied areas within the machine learning community. However, there has been a trend in applying meta-learning in novel and less-explored setups in recent years. It is worth noting that while data efficiency is a central consideration in FSL, time efficiency in terms of fast adaptation to the existing data within each task is critical in many of these settings, elucidating why meta-learning is broadly adopted in these novel scenarios. 

Meta-learning has been applied in various games such as zero-sum games, potential games, general sum multi-player games, and Stackelberg security games, showing significant improvements \citep{harris2022meta}. Meta-learning an optimal initialization has also been applied in combinatorial optimization \citep{wang2023unsupervised} by recasting unsupervised learning for combinatorial optimization as a meta-learning problem, as well as few-shot knowledge graph completion \citep{wu2022hierarchical}. Additionally, meta-learning has been extended to learning cognitive tasks \citep{miconi2023learning} by introducing another evolutionary level of learning that enables modification of the network in a self-referential manner. Finally, meta-learning has been used to discover optimizing functions without access to gradients efficiently \citep{lange2023discovering} and to solve optimal transport problems \citep{amos2023meta} rapidly.

\subsection{Green AI via FSL} 

A central characteristic of modern LLMs and foundation models is their scaling in the regime of billions of parameters. While this has led to breakthrough performance results, it has also led to a commensurate increase in their energy consumption and carbon emissions. The environmental damage associated with training and operating such models has been highlighted in 2023's AI index report \citep{maslej2023artificial}, showing that their carbon emissions exceed those of an average person's lifetime. Other studies \citep{patterson2021carbon} have also focused on measuring the carbon footprint of LLMs, yet more needs to be done to mitigate their impact.

A first step toward this direction has been the development of novel Green AI metrics that quantify these effects. For instance, in \cite{wu2022sustainable}, the carbon footprint of AI models is estimated as:

\begin{equation}
    CO_2^{embodied} = \sum_i \frac{Time}{Lifetime}CO_2^{embodied}(AI_{system})(i)
\end{equation}

\noindent where $Time$ refers to the time the model is utilized over the specified $Lifetime$. Other tools that enable measurement of model training environmental footprint have also been proposed. On the other hand, in \cite{henderson2020towards}, a formula for calculating the total energy consumption is derived:

\begin{equation}
    e_{total} = PUE \sum_p (p_{dram}e_{dram} + p_{cpu}e_{cpu} + p_{gpu}e_{gpu})
\end{equation}

\noindent where $p_x$ refers to the percentage of each system resource used by process $x$ relative to the total resources used, $e_x$ is the energy usage of that resource, and \textit{PUE} is the Power Usage Effectiveness index of the data center. However, a unified benchmark and suitable metrics to calculate an AI model's overall carbon footprint are still missing.

Approaches such as FL have been explored in the context of Green AI, yet they face limitations due to hardware constraints, scarcity of renewable energy sources for edge devices, and the need for communication among devices. However, when combined with FSL, they can reduce energy consumption on each device up to 41.2\% \citep{cai2023federated}. Overall, FSL can alleviate the need for extensive model selection and hyperparameter tuning and reduce the volume of available data needed to achieve strong generalization performance. However, more research is required to establish FSL as a feasible solution for Green AI.

\subsection{Meta-Learning vs Transfer Learning debate}

While meta-learning has been the leading paradigm for FSL for many years, various recent results have cast doubt on its dominance. Specifically, feature reuse has been identified as the main factor for fast adaptation in MAML \citep{raghu2019rapid} while recent findings show that finetuning with modern pretrained backbones can also form strong baselines \citep{el2021lessons}. Consequently, transfer learning has emerged as a capable competitor of meta-learning, with numerous simple baselines based on finetuning being proposed and demonstrating comparable performance against complex meta-learning architectures.

A shared insight in this line of work is that learning strong, transferable representations is crucial for FSL \citep{wang2022revisit}, and the quality of these representations can be affected by various factors such as model size \citep{tian2020rethinking}, invariance and equivariance \citep{rizve2021exploring}, support set size \citep{luo2023closer}, and number of pretraining classes \citep{dhillon2019baseline}. Overall, further research is required to understand more clearly the importance of expressive and generalizable representations in FSL. An interesting direction would also be to examine FSL's interplay with other learning paradigms with similar objectives, such as self-supervised learning \citep{ni2021close} and incorporate techniques from these fields to obtain more generalizable representations \citep{tian2020rethinking}.

Moreover, the lack of a unified evaluation protocol has constituted a major issue in comparing meta-learning and finetuning models. It is argued that progress using meta-learning techniques might be illusory, with incongruities in the implementation details of various FSL methods \citep{dhillon2019baseline}, and lack of evaluation under domain shift \citep{chen2019closer}, identified as two main challenges obstructing meaningful comparative assessments. To mitigate these issues, a hardness metric has been proposed \citep{dhillon2019baseline}, as well as the use of novel datasets \citep{dumoulin2021comparing}. However, more research is needed to determine more datasets and a unified evaluation protocol for FSL methods.

\subsection{Impact of underlying distributions}

One of the main differences between FSL and standard machine learning problems is the limited number of support data within each task, which complicates the precise estimation of the underlying data distribution. While meta-learning approaches compensate for it by extracting shared knowledge across different tasks, it is still unclear how to effectively achieve accurate estimates of the underlying task-generating process. Since FSL typically relies on utilizing a set of tasks, this setting introduces another distribution at the meta-level, specifically the distribution of tasks. Consequently, similar to how data augmentation has been extensively utilized in standard machine learning to obtain a better evaluation of the underlying data-generating distribution, similar techniques can also be utilized at the meta-level to increase across-task generalization by learning a more accurate representation of the underlying task-generating distribution. Towards that direction, in recent years, there has been substantial progress in methods that employ interpolation and augmentation techniques at the meta-level.

In the case of task augmentation, various techniques, including Conditional Batch Normalization \citep{sun2023metamodulation}, mixing support and query sets with Manifold Mixup to create new query sets \citep{yao2021improving}, and adversarial training \citep{wang2021cross} have been employed to generate new tasks. On the other hand, task augmentation techniques typically include leveraging information from two existing tasks to create a new one. Some of the most recent approaches include using memory banks \citep{hu2023learning}, interval bounds \citep{datta2023interval}, Manifold Mixup \citep{yao2021meta}, and expressive neural set functions \citep{lee2022set} to encode and interpolate between tasks.

Yet, despite these practical implementations, we still lack a more enhanced comprehension of the influence of the underlying data distributions on FSL at a theoretical level. Specifically, only a single work \citep{zuo2022understanding} that focuses on the theoretical examination of FSL emphasizes the importance of careful modeling of both the task distribution and the data distribution within each task. Consequently, further research is needed to understand more clearly the connection between task and data distributions and how they can efficiently be leveraged in the context of FSL.

\subsection{Adoption of FSL in new modalities}

Traditionally, most FSL and meta-learning techniques have commonly been applied in the visual domain \citep{snell2017prototypical} and for RL \citep{wang2016learning}. However, since FSL aims to alleviate the adaptation problem in cases where data is scarce, its application could be fruitful in various other domains and data modalities.

A common type of data underexplored in the FSL setting is tabular data. Only recently, some novel approaches have emerged, including using Conditional Batch Normalization to create novel tasks with tabular data \citep{sun2023metamodulation} and applying prototypical networks on tabular data tasks created via unsupervised meta-learning \citep{nam2023stunt}. On a broader setting focusing on small tabular tasks, in \cite{hollmann2023tabpfn}, the authors have pretrained a transformer-based model on synthetically generated tasks using Bayesian Neural Networks and Structural Causal Models and showed that the model can adapt to downstream tabular tasks without any additional finetuning, similar to the way LLMs perform ICL. Time series data have also received insufficient attention until recently. Some notable first attempts in applying FSL in that modality include a meta-learning approach for few-shot forecasting in high-dimensional time series based on a single latent dynamic function conditioned in each task \citep{jiang2022sequential}, load forecasting via MAML-based models \citep{xu2022automated}, and meta-learning neural processes with application on point processes \citep{bae2023meta}.

Overall, one of the main reasons hindering FSL progress outside the visual domain and RL is the limited number of FSL datasets for these domains and modalities. Although some novel FSL datasets, such as FS-MOL \citep{stanley2021fs} for few-shot drug discovery, have emerged, the lack of similar datasets in modalities such as time series necessitates developing and establishing standardized FSL benchmarks in these domains.

\subsection{Fully generalizable FSL} 

The increasing popularity of foundation models on vision and language tasks has also given rise to various multimodal foundation models \citep{radford2021learning}. While many of these models demonstrate strong FSL performance or have been adapted to be effective in that setting, their shortcomings in achieving holistic and human-like generalization capabilities are still notable. Broadly speaking, generalization can be multi-faceted and may be abstracted in different levels based on context:

\begin{enumerate}
    \item \textbf{Single instance of a learning problem.} Better known as domain adaptation/generalization, the model's goal is to generalize strongly to data from a previously unseen distribution for a specific problem.

    \item \textbf{Multiple instances of the same learning problem. } Generalization at that level could be seen as being able to adapt to different variations of the specific problem, e.g., multi-class classification with a varied number of classes and labels.

    \item \textbf{Multiple learning problems of the same modality. } The goal at this level is to effectively generalize to different learning problems of the same modality, e.g., from image classification to object detection.
 
    \item \textbf{Multiple domains. } At the highest level of abstraction, a model would be able to generalize across different modalities even if the model has never been trained explicitly on them.

\end{enumerate}

While the first two cases have been extensively researched for FSL, only recently has generalization within a specific modality been explored via models that can generalize across different visual prediction tasks \citep{kim2023universal, bachmann20244m}, and the curation of FSL datasets containing multiple visual learning problems \citep{bohdal2023meta}. However, current vision learning models struggle to generalize to new learning problems, and overall, FSL methods still face challenges in extrapolating to unseen modalities. Creating a genuine generalist model capable of FSL across novel modalities could be an ambitious but exciting future direction.

\subsection{Human-like learning}

One of the major attributes of human intelligence is the ability to learn new concepts quickly, master new tasks, and combine them to solve complex problems, given only a few examples, possibly due to humans' inherent inductive biases and their ability to leverage multimodal signals \citep{lin2023multimodality}. However, current machine learning models struggle with fast adaptation to complex tasks and lack human-like intelligence attributes such as compositionality, concept association, relational understanding, and conceptual meaning \citep{jiang2023mewl}. Systematicity also remains a challenge for modern models \citep{lake2023human}.

Rapid adaptation is generally prevalent in nature and has been studied for many decades in cognitive science, psychology, and neuroscience, with results highlighting the existence of different and nested timescales of learning. These observations have also been utilized in meta-learning via hierarchical Bayesian models \citep{lake2015human}. The compositionality and expressiveness of natural language have also been leveraged for FSL of concepts, focusing on induction and utilizing a human-like prior extracted from behavioral data from humans \citep{ellis2023modeling}. Moreover, meta-learning has been used for human-like learning of algebraic systems, leading to models that learn how to generalize and adopt human inductive biases \citep{lake2023human} systematically. On a different direction, brain mechanisms such as synapse formation in the human brain have inspired novel approaches in FSL, such as adaptation mechanisms for few-shot generative domain adaptation, enhancing the model with human-like properties such as memory and domain association \citep{wu2023domain}. Lastly, it has been shown that increased AI alignment can improve FSL models, adaptation to new domains, and robustness to adversarial examples, possibly by serving as a source of inductive bias that reduces the amount of information needed from data \citep{sucholutsky2023alignment}.

It is worth noting that while cognitive science and neuroscience have focused on discovering the mechanisms of human intelligence, machine learning focuses on replicating these capabilities. In that direction, a new multimodal dataset for FSL has been proposed that enables few-shot multimodal abstract reasoning evaluation of models over various cognitive tasks \citep{jiang2023mewl}. Preliminary results on the dataset suggest significant misalignment between machines and humans regarding FSL capabilities, raising questions on how to align machine and human learning and create agents with human-like FSL capabilities. Consequently, methods combining neuroscience and cognitive science with machine learning might be the way forward.

\section{Conclusion} \label{conclusion}

Although deep learning has traditionally relied on large amounts of data to facilitate learning, there is an increasing focus on techniques that can learn with limited data. Similarly, despite the trend of creating large foundation models trained on massive datasets, there is a growing interest in exploring new FSL techniques and applying them in practical settings. Given FSL's recent rapid expansion, this paper aims to provide an overview of this research field, covering its various aspects. The standard FSL formulation is initially provided, based on which different proposed methods are developed. To position FSL within the broader machine learning paradigm, its connection with different learning fields is also examined. Subsequently, we present a comprehensive taxonomy of FSL methods that extends existing taxonomies with recent approaches and model families and also introduces emerging learning paradigms within FSL, such as in-context learning. Additionally, we provide a broad overview of hybrid approaches that extend FSL beyond the presented supervised learning setting. FSL's bloom, which is also reflected in the numerous real-world applications in which it has been applied, is also described. Finally, we discuss various emerging trends, including the few-shot evaluation of foundation models and the application of meta-learning in novel settings, and aim to encourage further research by identifying key areas where FSL can have a significant impact, such as Green AI and bridging the gap between machine learning and human intelligence.

\bmhead{Acknowledgements}

This project has received funding from the European Union’s Horizon Europe research and innovation programme under grant agreement No. 101135800 (RAIDO).

\bibliography{bibliography}

\end{document}